\documentclass{article}

\PassOptionsToPackage{numbers, compress}{natbib}

\usepackage[final]{neurips_2019_ml4ad}

\usepackage[utf8]{inputenc} 
\usepackage[T1]{fontenc}    
\usepackage{hyperref}       
\usepackage{url}            
\usepackage{booktabs}       
\usepackage{amsfonts}       
\usepackage{nicefrac}       
\usepackage{microtype}      
\usepackage{graphicx}
\usepackage{caption}
\usepackage{subcaption}
\usepackage{todonotes}
\usepackage{caption}
\usepackage{subcaption}
\newcommand*{\addheight}[2][.5ex]{%
  \raisebox{0pt}[\dimexpr\height+(#1)]{#2}%
}
\title{Joint Interaction and Trajectory Prediction for Autonomous Driving using Graph Neural Networks}

\author{%
    Donsuk Lee \\
    School of Informatics, Computing, and Engineering\\
    Indiana University, Bloomington, IN\\
    \texttt{donslee@iu.edu} \\
    \And
    Yiming Gu \\
    Uber ATG \\
    50 33rd St, Pittsburgh, PA\\
    \texttt{yiming@uber.com} \\
    \AND
    Jerrick Hoang \\
    Uber ATG \\
    50 33rd St, Pittsburgh, PA\\
    \texttt{jhoang@uber.com} \\
    \And
    Micol Marchetti-Bowick \\
    Uber ATG \\
    50 33rd St, Pittsburgh, PA\\
    \texttt{mmarchettibowick@uber.com} \\
}

\begin{document}

\maketitle

\begin{abstract}
In this work, we aim to predict the future motion of vehicles in a traffic scene by explicitly modeling their pairwise interactions. Specifically, we propose a graph neural network that jointly predicts the discrete interaction modes and $5$-second future trajectories for all agents in the scene. Our model infers an interaction graph whose nodes are agents and whose edges capture the long-term interaction intents among the agents. In order to train the model to recognize known modes of interaction, we introduce an auto-labeling function to generate ground truth interaction labels. Using a large-scale real-world driving dataset
, we demonstrate that jointly predicting the trajectories along with the explicit interaction types leads to significantly lower 
trajectory error than baseline methods. Finally, we show through simulation studies that the learned interaction modes are semantically meaningful.
\end{abstract}

\section{Introduction}

Developing autonomous vehicles that can drive on public roads along with human drivers, pedestrians, cyclists, and other road users is a challenging task. Researchers have been attempting to solve this problem for many years, from the days of ALVINN \cite{pomerleau1989alvinn} and the DARPA Urban Challenge \cite{montemerlo2008junior, urmson2008autonomous} to exploring a variety of approaches in recent years, including end-to-end learning \cite{chen2015deepdriving} and traditional engineering stacks \cite{levinson2011towards}. In order to drive both safely and comfortably in the real world, one of the most important and difficult tasks for self-driving vehicles (SDVs) is to predict the future behaviors of the surrounding road users.

\begin{figure}
\centering
\begin{subfigure}{.6\textwidth}
  \centering
  \includegraphics[width=.9\linewidth,trim={0 3cm 0 0cm},clip]{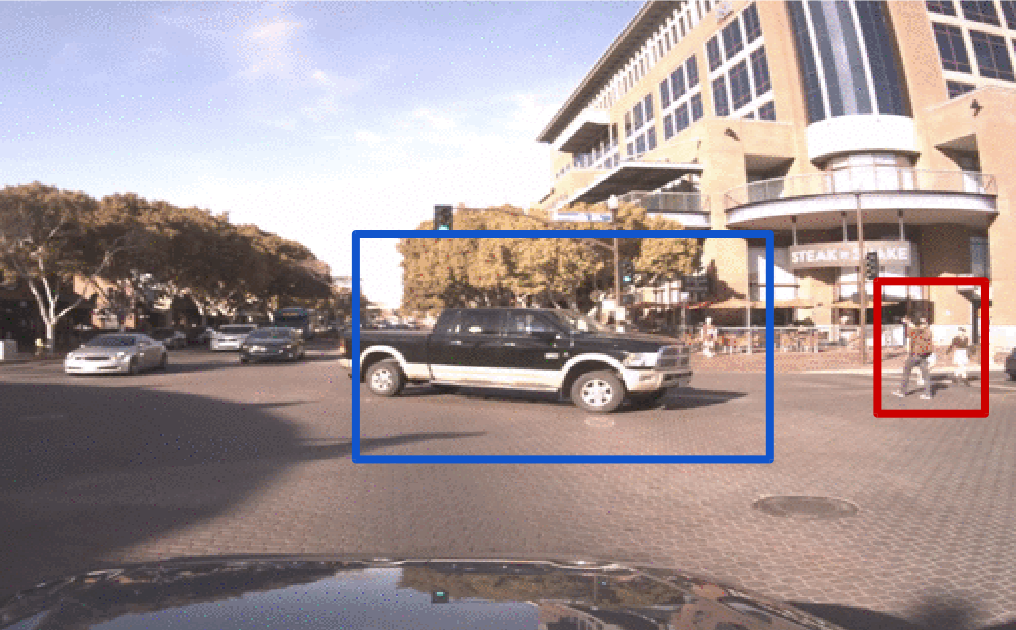}
  \caption{Real image}
  \label{fig:motivatingexamplesub1}
\end{subfigure}%
\begin{subfigure}{.4\textwidth}
  \centering
  \includegraphics[width=.9\linewidth,trim={0 0 0 4cm},clip]{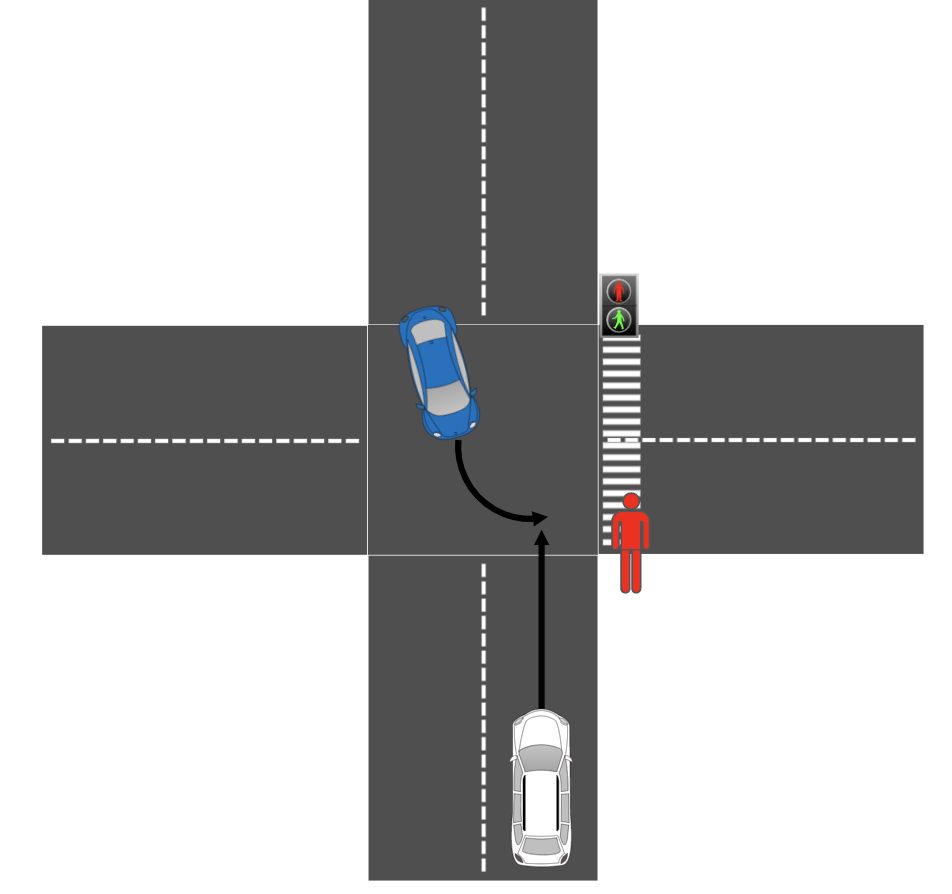}
  \caption{Cartoon illustration}
  \label{fig:motivatingexamplesub2}
\end{subfigure}
\caption{A traffic scene in which an SDV (white) approaches an intersection where another vehicle (blue) is interacting with a pedestrian (red). Failing to capture the interaction between the vehicle and the pedestrian when predicting their future behavior may lead the SDV to execute a poor motion plan.}
\vspace{-.2cm}
\label{fig:motivatingexample}
\end{figure}

There has been significant research dedicated to predicting the future states of traffic actors. One common line of research attempts to independently predict each actor's future trajectory from the scene \cite{altche2017lstm, cui2019multimodal, deo2018multi, kim2017probabilistic, luo2018fast, xie2017vehicle}. However, a limitation of many of these approaches is that they fail to capture the interactions among actors. For example, in the case shown in Figure~\ref{fig:motivatingexample}, the future trajectory of the blue vehicle and the future trajectory of the pedestrian depend on one another. In this case, there are two possible outcomes of the interaction: the pedestrian yields to the vehicle, or, more likely, the vehicle yields to the pedestrian. Predicting the marginal future trajectory of the vehicle or the pedestrian will ignore the possible modes of interaction between the two actors, and will end up capturing an inaccurate distribution over the future trajectories.

Recently, there has been an increasing amount of work on modeling interaction in multi-agent systems with neural networks \cite{alahi2016, battaglia2018, deo2018, hoshen2017, kipf2018, rhinehart2019precog, sukhbaatar2016learning, sun2018, tacchetti2018, casas2019spatially}. For example, in CommNet~\cite{sukhbaatar2016learning}, communication protocols (interactions) between smart agents are learned in conjunction with the final prediction outcomes through a Graph Neural Network (GNN), where the communication is learned implicitly. Similarly, in SocialLSTM~\cite{alahi2016}, interaction between pedestrians is captured by the \emph{social pooling} operation on the hidden states of the LSTMs. As opposed to modeling interaction implicitly, Neural Relational Inference~\cite{kipf2018} models the interactions in dynamical systems as latent edge types of an interaction graph, which are learned in an unsupervised manner.

Another related approach in the autonomous driving domain is IntentNet~\cite{casas2018intentnet}. In this work, the model learns discrete actions, such as ``keep lane'' and ``left lane change'' using supervision. One limitation of predicting actions instead of interactions is that it is unnatural to pose constraints or priors on a pair of actor actions, but much easier to do so with interactions. An example of such a prior is illustrated in Figure~\ref{fig:motivatingexample}, where we believe that if the pedestrian goes first, then the vehicle will yield, and vice versa. By introducing the concept of pair-wise interaction, we are able to capture the fact that in this interaction pair, it is unlikely that both the vehicle and the pedestrian will go at the same time and it is instead more likely that one actor will yield to the other. Importantly, by using the future observations of the scenario to categorize interaction types during training, we can learn these pair-wise interactions, instead of independent agent-wise actions, \emph{explicitly} in a supervised manner. 

In this paper, we propose a supervised learning framework for the joint prediction of interactions and trajectories. Specifically, we model interactions as intermediate discrete variables that capture the long-term relative intents of the actors, such as whether one actor will yield to another. 
In order to learn the interaction types from labeled examples, we introduce a labeling function which uses simple heuristics to programmatically generate the labels from the future trajectories. This enables us to build a large dataset of vehicle interactions without relying on human experts for manual labeling. In addition to improving the accuracy of trajectory predictions, we show that explicitly modeling the interaction types helps capture the modes of vehicles' future behaviors in an explainable manner. Our approach is empirically verified by experiments conducted on a large-scale dataset collected by real-world autonomous vehicles.

\section{Problem Formulation}

We are mainly interested in predicting the future dynamics of multi-agent systems consisting of vehicles. Our goal is to predict the future trajectories of vehicles in traffic scenes, given their observed states and some additional features describing the traffic conditions around them. In order to jointly model the dynamics of all agents in the system in a structured way, we introduce an auxiliary task of learning discrete interaction types between agents.

We first define the state $\mathbf{s}^i_t$ to be the 2D position and velocity of agent $i$ at time $t$, and let $\mathbf{s}_{t_1:t_2}^i$ denote the sequence of agent $i$'s states from $t_1$ to $t_2$. For compactness, we further define ${S}_{t_1:t_2}=\{\mathbf{s}_{t_1:t_2}^i\}_{i=1:N}$ to be the state sequences of all $N$ agents in the scene. Then, the trajectory prediction task is to predict the future states ${S}_{t:t+T}$ of all agents given observations of their past states ${S}_{t-T:t}$.

Next, we assume that there exist discrete types that summarize the modes of interaction between each pair of agents. 
Under this assumption, we introduce a secondary task of learning the interaction types from labeled examples. In traffic scenarios, it is often difficult to capture interactions based solely on the agents' dynamics, and additional contextual features about the agents in the scene can be very informative. Let $\mathbf{a}_i \in \mathbb{R}^{F_a}$ and $\mathbf{p}_{ij} \in \mathbb{R}^{F_p}$ be agent-wise and pair-wise features that describe an individual agent (capturing basic traffic context) and the relationship between a pair of agents (capturing their relative dynamics via a compact set of high-level features), respectively. Then, the interaction prediction task is to predict the interaction label $l_{ij}$ for each ordered pair of agents $(i, j)$ given the agent-wise and pair-wise features along with the observed dynamics of the pair. 

Since our primary goal is still trajectory prediction, we combine the predicted interaction types $\hat{l_{ij}}$ with information about the agents' past states and provide these as inputs to the trajectory prediction module. Explicitly capturing these interaction types guides the trajectory prediction module on how to aggregate information from agents $j \neq i$ when predicting the future behavior of agent $i$, which ultimately leads to more accurate trajectories.


\section{Method}
\label{sec:method}

We tackle the trajectory prediction problem by jointly learning to predict both interaction types and future trajectories. The key insight is that by learning interaction types and future dynamics jointly, a model can learn to make better and more explainable predictions.

\textbf{Labeling Function.}
Supervised learning of interaction types requires labeled examples. Instead of obtaining interaction labels from human experts, we use simple heuristics to  programmatically generate the labels, similar in philosophy to \cite{zhan2018}. We extract an interaction label $l_{ij}$ for each ordered pair of agents $(i, j)$ in the scene at each timestep $t$. The label is determined by the future trajectories of the agents. 
Given the trajectories, the labeling function outputs: a) $l_{ij} = \texttt{IGNORING}$ if trajectories do not intersect, b) $l_{ij} = \texttt{GOING}$ if trajectories intersect and $i$ arrives at the intersection point \textit{before} $j$, and c) $l_{ij} = \texttt{YIELDING}$ if trajectories intersect and $i$ arrives at the intersection point \textit{after} $j$.

\textbf{Graph Representation of Agent States.}
It is undesirable to use a global coordinate system in trajectory prediction tasks because of the high variability of the input and output coordinates \cite{becker2018}. Instead, we transform the trajectory waypoints of each agent to their individual reference frame. The current position of an agent at time $t$ is set to be the origin of the agent's reference frame, and the coordinates of past and future trajectory points are offset by the agent's current position and rotated by the current heading of the agent in a global frame.

When the coordinates of the past and current states of each agent are transformed from global frame to that individual agent's coordinate frame, information about the relative positions and velocities among agents is lost. However, a model needs to be informed with the relative configuration of agents in order to reason about their interaction. To preserve the relationships among agents, we represent the configuration of agents at a given timestep as a state graph, in which a node represents the state of an individual agent in its reference frame, and a directed edge represents the relative state of the destination node in the source node's reference frame. 

\textbf{Modeling Interaction with Graph Network.}
We use a variant of Graph Network (GN) layers \cite{battaglia2018} to process the state graphs and model the interactions between agents. Our GN consists of two components: an edge model which combines the representations of each edge and its terminal nodes to output an updated edge representation, and a node model which operates on each node to aggregate the representations of incident edges and outputs an updated node representation.
We model different types of interaction using a separate learnable function $f_m$ for each type $m$ in the edge model (see Appendix for details). Given the predicted scores of the interaction types between a pair of actors, the edge model computes the sum of the outputs from each $f_m$ weighted by these scores.

\begin{figure}
\centering
\includegraphics[width=.85\linewidth]{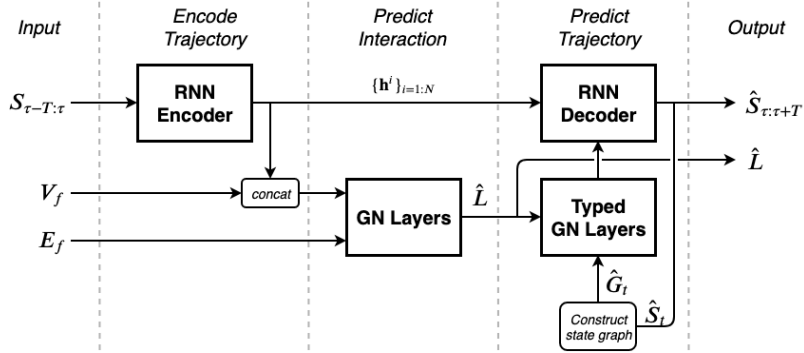}
\caption{A joint model for interaction and trajectory prediction. See text in Section~\ref{sec:method} for details.}
\label{fig:jointmodel}
\vspace{-.1cm}
\end{figure}

Figure~\ref{fig:jointmodel} describes the overall schematics of our joint prediction model. Our model consists of three components: 1) trajectory encoder network, 2) interaction prediction network, and 3) trajectory decoder network. First, the encoder network encodes the observed past states $\mathbf{s}^i_{\tau-T:\tau}$ of each agent into a hidden state $h^i$. Then, the interaction prediction network takes in the encoded states of all agents $H = \{h^i\}_{i=1:N}$, along with the agent-wise features $V_f = \{\mathbf{a}_i\}_{i=1:N}$ and pair-wise features $E_f = \{\mathbf{p}_{ij}\}_{i,j=1:N, j=1:N, i \neq j}$, and predicts the interaction type scores $\hat{L}=\{\hat{l}_{ij}\}_{i=1:N, j=1:N, i \neq j}$ for every ordered pair of agents, using a stack of two vanilla (untyped) GN layers. 
Finally, given the hidden states $H$, interaction type scores $\hat{L}$, and the initial states $S_{\tau}$, the decoder network rolls out the future states of the agents $\hat{S}_{\tau:\tau+T}$. This module aggregates information from all actors using a stack of two typed GN layers, which employ an MLP for each learned function $f_m$ in the edge model.

\textbf{Loss Function.} The loss function we use is the combination of a classification loss over the edges for predicting the discrete interaction types (edge loss) and a regression loss over the nodes for predicting the continuous future trajectories (node loss). Our complete loss function is given by $$L = \alpha ~ \textrm{CE}(L, \hat{L}) + \textrm{MSE}(S_{\tau:\tau+T},\hat{S}_{\tau:\tau+T}) $$ 
where $\textrm{CE}()$ is the Cross-Entropy loss and $\textrm{MSE}()$ is the Mean-Squared-Error loss. For the supervised interaction model, we set $\alpha=5$, and for the unsupervised interaction model, we set $\alpha=0$. 

\section{Experiments}
We present experiments on real-world traffic data collected by autonomous vehicles which operated in numerous locations across North America. The dataset contains trajectories of 68,878 unique vehicles in various traffic scenarios. The vehicles were tracked continuously with a sampling frequency of 2 Hz (0.5s interval). We sample trajectories in sliding time windows of 10 seconds, and use the first 5 seconds as inputs and the last 5 seconds as prediction targets. Using the extracted trajectories, we run our labeling function to obtain the labels for every pair of agents that are less than 100 meters apart from each other. To evaluate the performance of our models, we report the mean displacement error, cross-track error, and along-track error between the estimated and ground-truth trajectories. 
We present an ablation study to analyze the capability of our proposed method to capture interactions between agents. 
The quantitative results are summarized in Table~\ref{table:results}.

\begin{table}
\centering
\begin{tabular}[t]{lcccc}
\toprule
Method & mean DPE & mean ATE & mean CTE\\
\midrule
Baseline, no interaction & 2.051 &	1.818& 0.558\\
Graph, untyped, yielding/going edges only & 1.725	&1.511&	0.512\\
Graph, untyped, all edges	& 1.713 &1.491&	0.523\\
Graph, oracle, yielding/going edges only & 1.709 &	1.435	&0.519\\
Graph, oracle, all edges & 1.638 &	1.435&	0.489\\
Graph, joint, supervised interaction & 1.611 &	1.397&	0.500\\
Graph, joint, unsupervised interaction & \textbf{1.579} &	\textbf{1.378}&	0.477\\
Trajectory prediction with map and scene context \cite{djuric2018motion} & 1.643 &	1.533&	\textbf{0.334} \\
\bottomrule
\end{tabular}
\vspace{.2cm}
\caption{Quantitative evaluation. We report the displacement error (DPE), along-track error (ATE), and cross-track error (CTE) in meters averaged over a $5$-second horizon.}
\label{table:results}
\vspace{-0.3cm}
\end{table}%

Our baseline model is an RNN encoder and decoder, which treats trajectories independently without modeling interactions between agents. In addition, we introduce two variants of our joint model. The first variant (\emph{untyped}) has a single edge function to learn interaction without differentiating between types. The second variant (\emph{oracle}) is modified to use ground-truth interaction types, instead of its own predictions, to predict trajectories. We also modify each variant to exclude the edges with $\texttt{IGNORING}$ labels from the graph in order to see if these edges can be ignored as the name suggests.

We first demonstrate the power of graph networks to model interaction by comparing our joint model and its variants against the baseline. We observe that all of the graph models significantly outperform the baseline. This suggests that the motion of vehicles is highly interdependent, and graph models can effectively capture their interactions. Next, we showcase the effect of interaction labels on trajectory prediction. The typed variants outperform all of the untyped variants, which suggests that our graph model benefits from the discrete modeling of interaction types. Furthermore, we can see that the typed model benefits from having information shared along the \texttt{IGNORING} edges. 



Finally, we present the results of our full fledged joint prediction model. Even without rich map context, our model shows comparable performance with~\cite{djuric2018motion}, particularly on along-track error, which captures the temporal accuracy of the predicted trajectories. Notably, another version of the joint model trained without supervision on interaction labels (simply by zeroing out the interaction classification loss) achieves better performance than the supervised model. This implies that the heuristics used in our labeling function are not optimal, and could be improved for better trajectory prediction. Nonetheless, we observe via simulation experiments that the supervised model predicts trajectories that are consistent with the meanings of the interaction labels (see Appendix and supplementary video for details). This interpretability helps provide key insights into the model's behavior, which is a crucial step towards building safe prediction systems for autonomous vehicles.

\section{Conclusion}
In this paper, we propose a graph-based model for multi-agent trajectory and interaction prediction, which explicitly models discrete interaction types using programmatically generated weak labels and typed edge models. The main advantages of our approach are:
i) we can gain a boost in performance without additional labeling costs when compared to the baseline, and ii) our model can effectively capture the multi-modal behavior of interacting agents while learning semantically meaningful interaction modes.





\bibliographystyle{abbrv}
\bibliography{neurips_2019_ml4ad_yiming}

\section{Appendix}

\subsection{Qualitative Evaluation on Simulated Data}

In order to understand the influence of the edge types on the final trajectory predictions generated by our model, we simulated several very simple dynamic interactions between two actors. We generated simulated historical states for the actors and then injected fixed values for the edge scores to analyze how the predicted trajectories change as a function of the injected edge types. We visualized the predicted trajectories from the baseline model, oracle model with all edges, supervised joint model, and unsupervised joint model while injecting several different combinations of interaction types. We compiled the results of these simulations into a short video, which is available at \url{https://www.youtube.com/watch?v=n5RNRDdoPoQ}. 

In these simulations, we find that all of the graph models learn to rely heavily on the edge type in order to predict the future trajectories of the two actors. We can effectively control how the interaction plays out simply by injecting different edge types (e.g.,~yielding/going vs.~going/yielding). The interaction modes of the oracle and supervised models correspond directly to the labeled categories that we provide. Interestingly, the interaction modes learned by the unsupervised model encode a similar set of categories, but seem to capture the leading/following relationship separately from the going/yielding relationship.

\subsection{Qualitative Evaluation on Real Data}

Next, we looked at some examples of real-world scenes and qualitatively evaluated our model's performance on these cases.  Table~\ref{table:trajvis1} illustrates some examples of trajectory predictions for actors driving in three-way and four-way intersections. The trajectories are predicted from time $t$ to $t+H$, where $H$ is the prediction horizon. The different colors indicate trajectory predictions for different actors. Each dot shows a single predicted waypoint (at 0.5-second intervals), and the more transparent the dot is, the further away it is in the future (i.e.,~the further its timestamp is from the current time, $t$). 

We observe several interesting patterns in these examples. First, note that the map in the figures is purely for illustration -- we do not provide map information directly to the model. Nevertheless, the graph models are able to learn the lane directions and drivable surfaces to some degree by observing the histories of the other vehicles. Second, we notice that the model that uses only the $\texttt{YIELDING}$/$\texttt{GOING}$ edges (second column) is substantially worse at capturing lane-following behavior than all other models (see rows (a), (b), (c), and (e) for examples), suggesting that the \texttt{IGNORING} edges are useful for transmitting implicit map information from actor to actor. 
Third, if we compare the supervised and unsupervised models (last two columns), we observe that the unsupervised model is slightly worse at predicting lane-following behavior than the supervised model (see rows (c), (d), and (e) for examples). We also notice that the supervised model appears to predict fewer conflicts between trajectories than the unsupervised model (an example can be seen in row (b), where the supervised model clearly predicts the yellow actor to yield to the blue actor, but the unsupervised model predicts that both will go at the same time). 
Lastly, we see that in case (e), the red merging actor may be equally likely to turn left or right, and because the current model is uni-modal (i.e., it only predicts the single most likely future trajectory), it is not able to model such discrete modes. This suggests two future works: (1) incorporating map elements (traffic signals, traffic signs, lane segments, sidewalks) as nodes in the graph; (2) adding multi-modality to the model.

\begin{table}[ht]
\setlength{\tabcolsep}{0.3pt}
\begin{tabular}{|c|c|c|c|c|c|}
      \hline
      \addheight{(a)} &
      \addheight{\includegraphics[trim=300 300 300 300, clip,width=28mm]{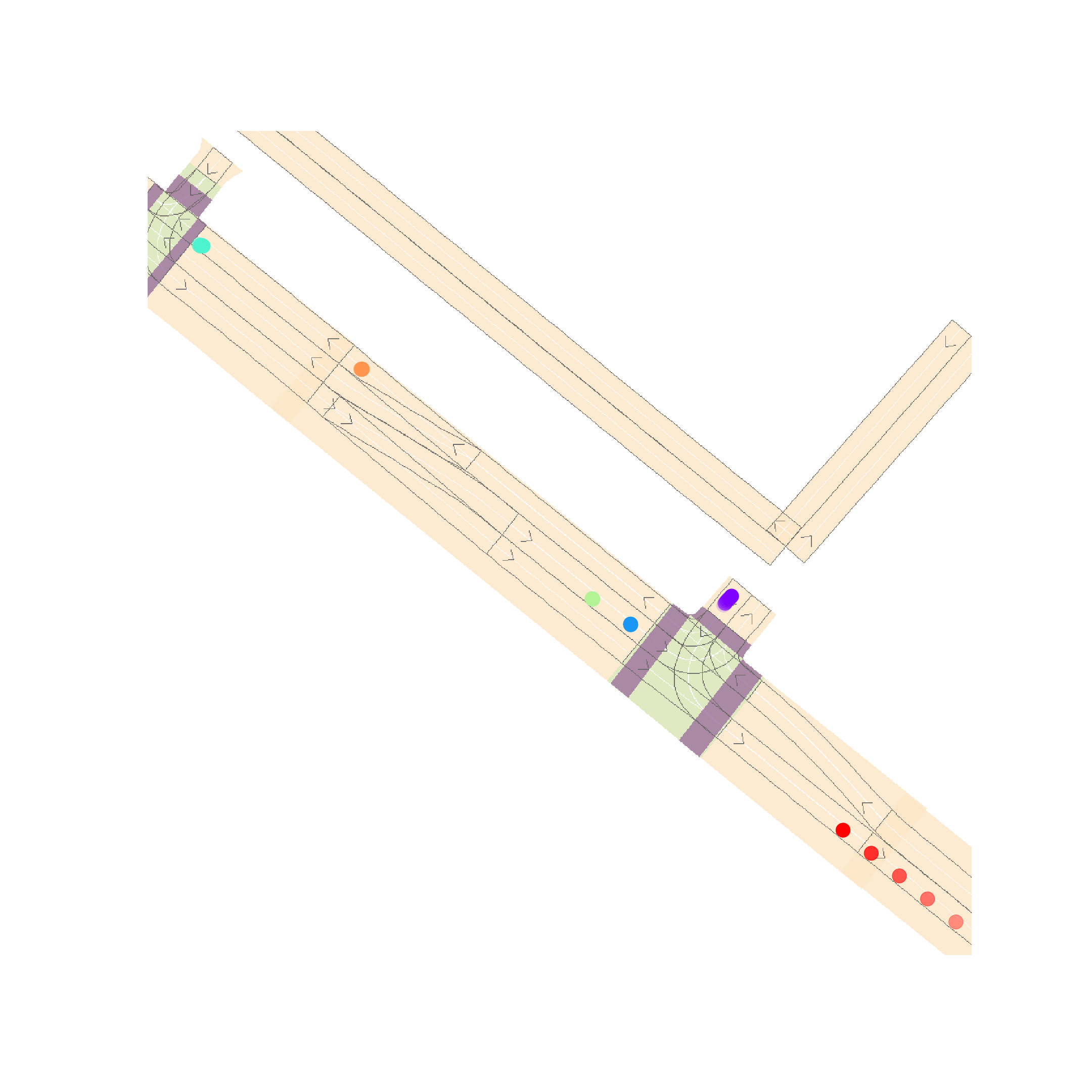}} &
      \addheight{\includegraphics[trim=300 300 300 300, clip,width=28mm]{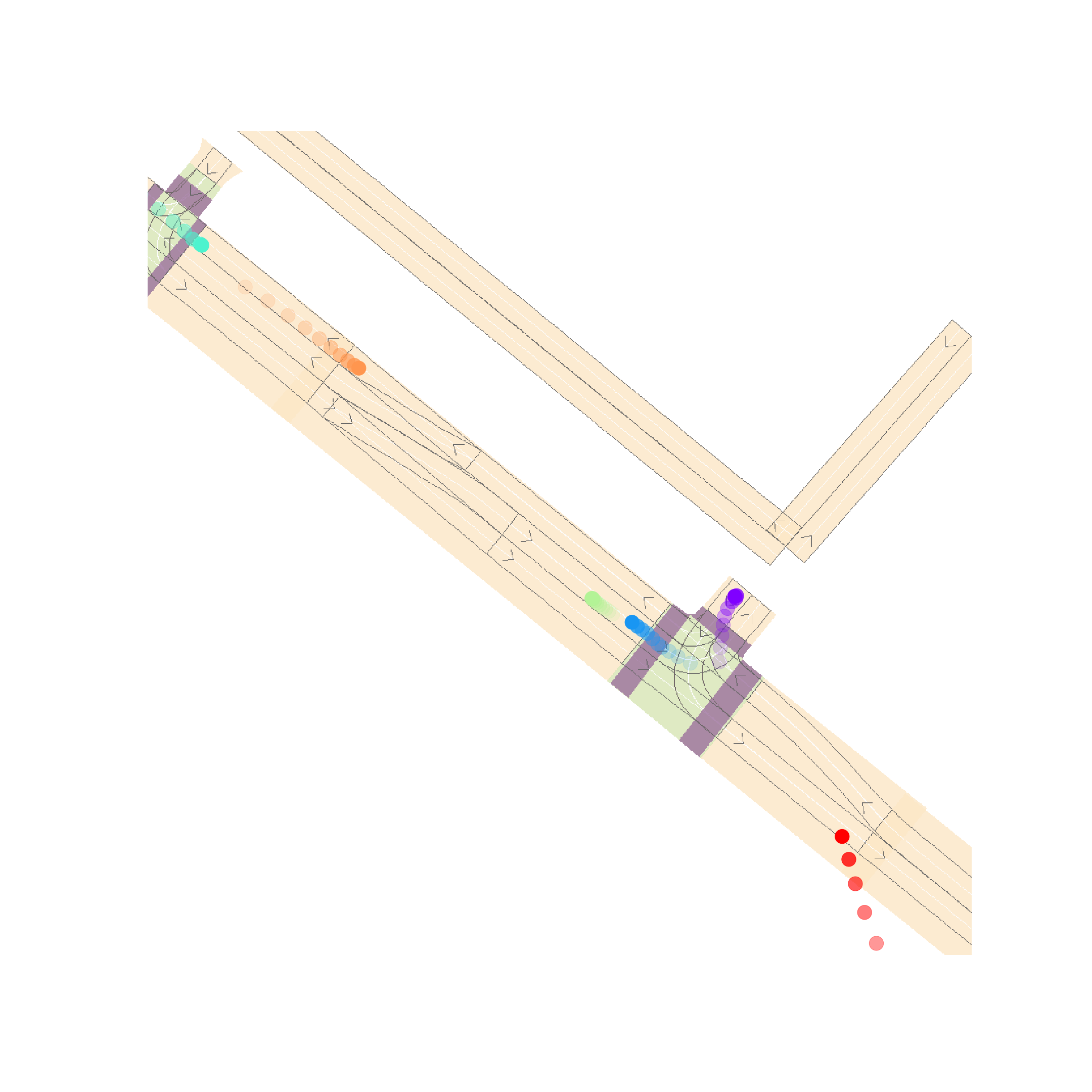}}&
      \addheight{\includegraphics[trim=300 300 300 300,clip, width=28mm]{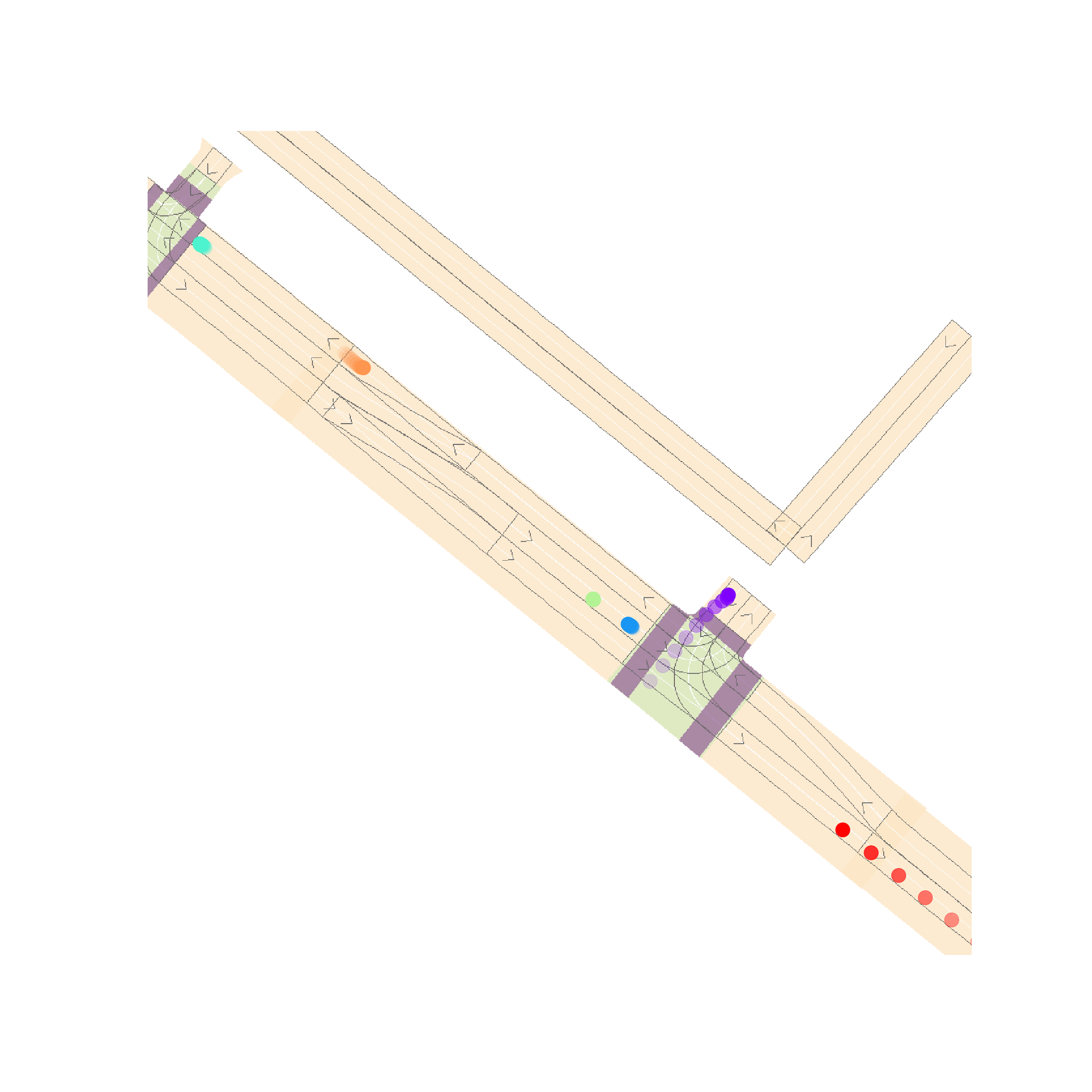}} & 
      \addheight{\includegraphics[trim=300 300 300 300,clip, width=28mm]{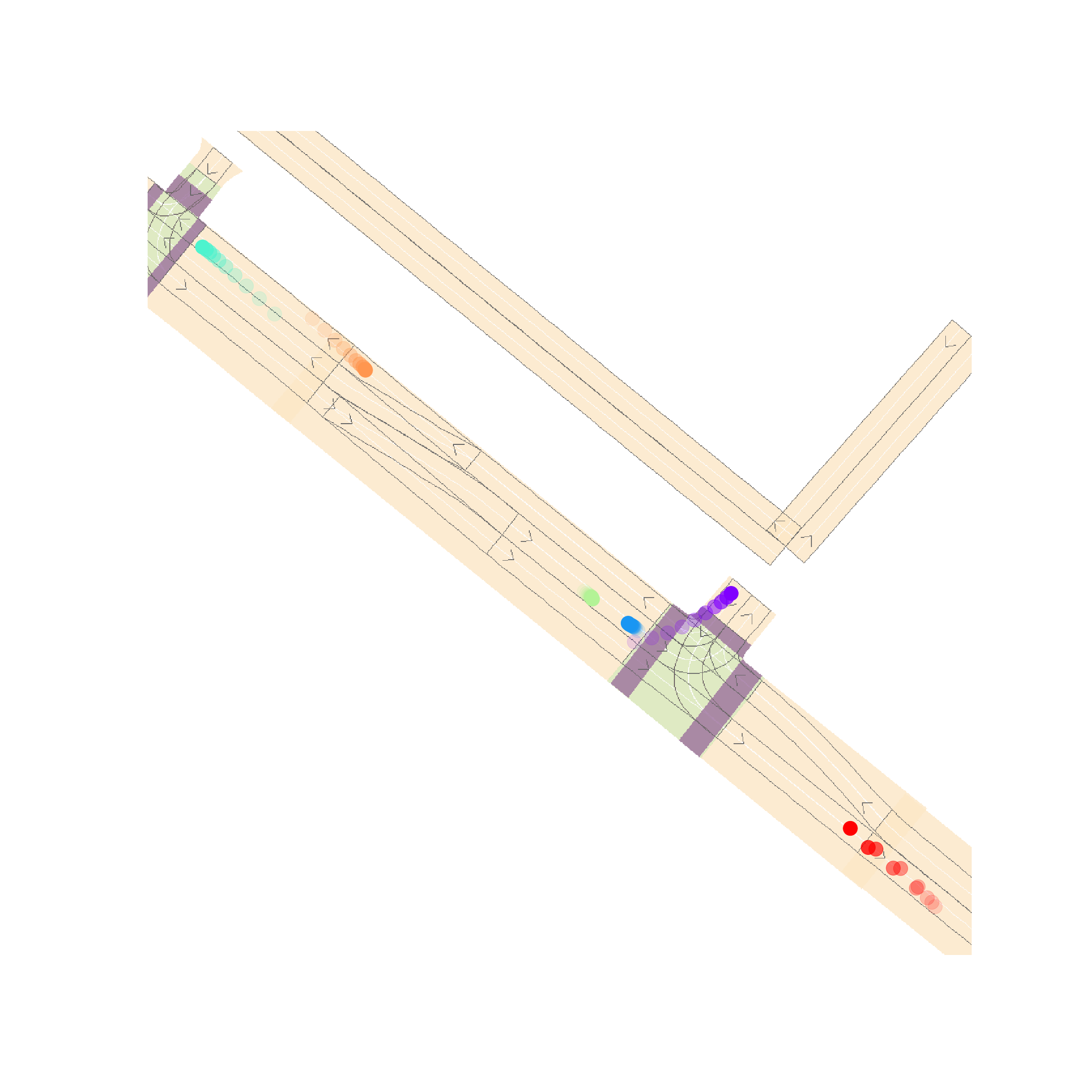}} &
      \addheight{\includegraphics[trim=300 300 300 300, clip,width=28mm]{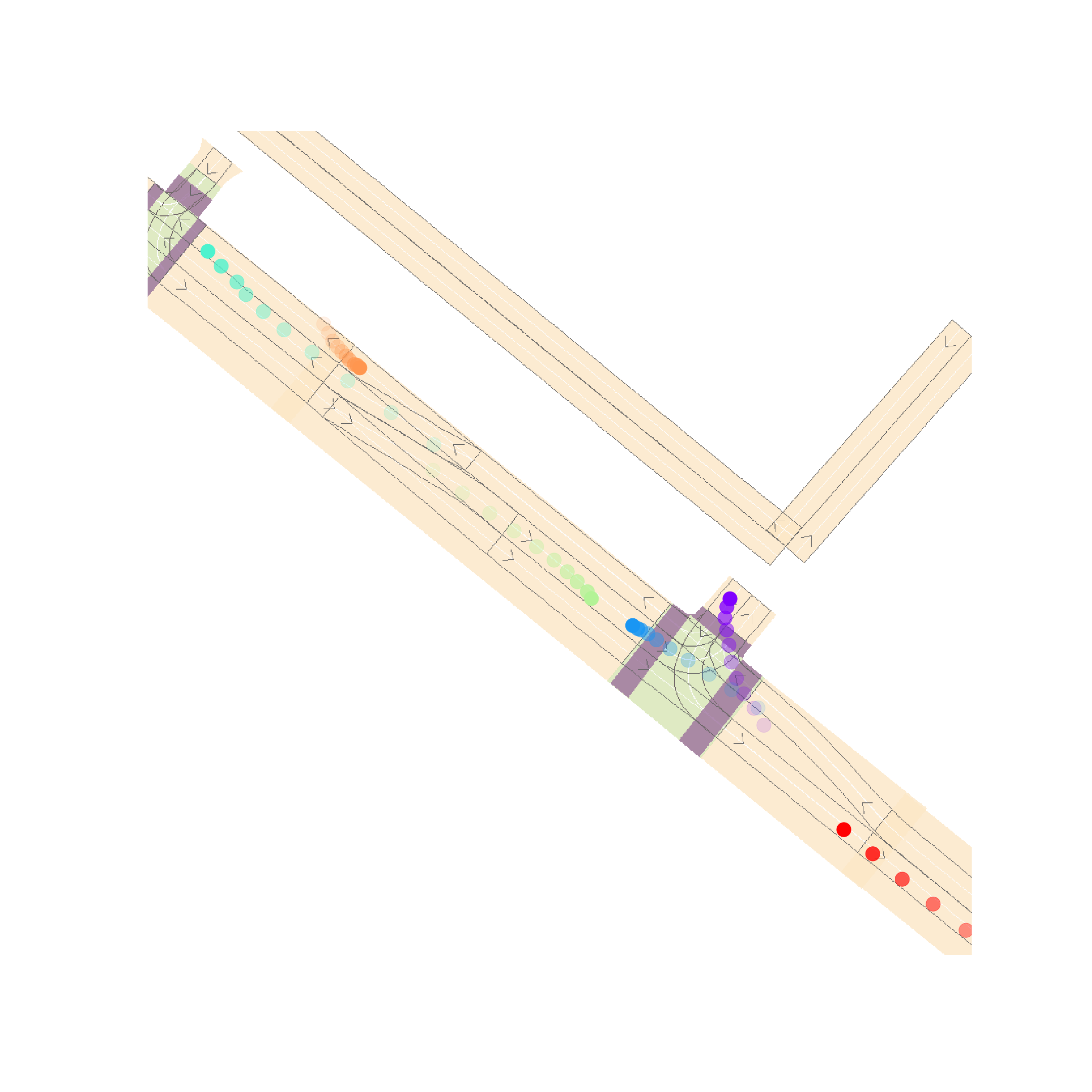}}\\
      \hline
      \addheight{(b)} &
      \addheight{\includegraphics[trim=300 300 300 300, clip,width=28mm]{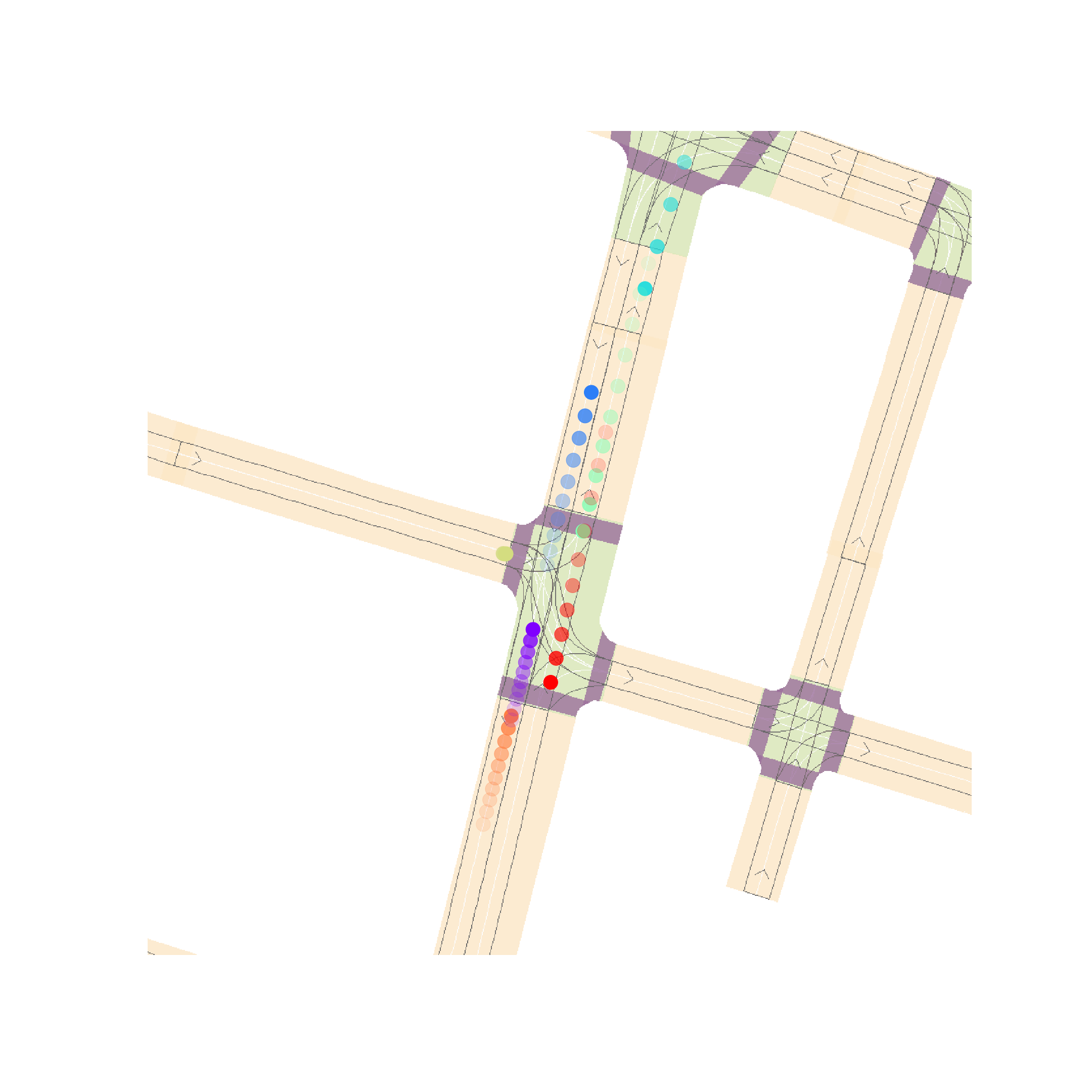}} &
      \addheight{\includegraphics[trim=300 300 300 300, clip,width=28mm]{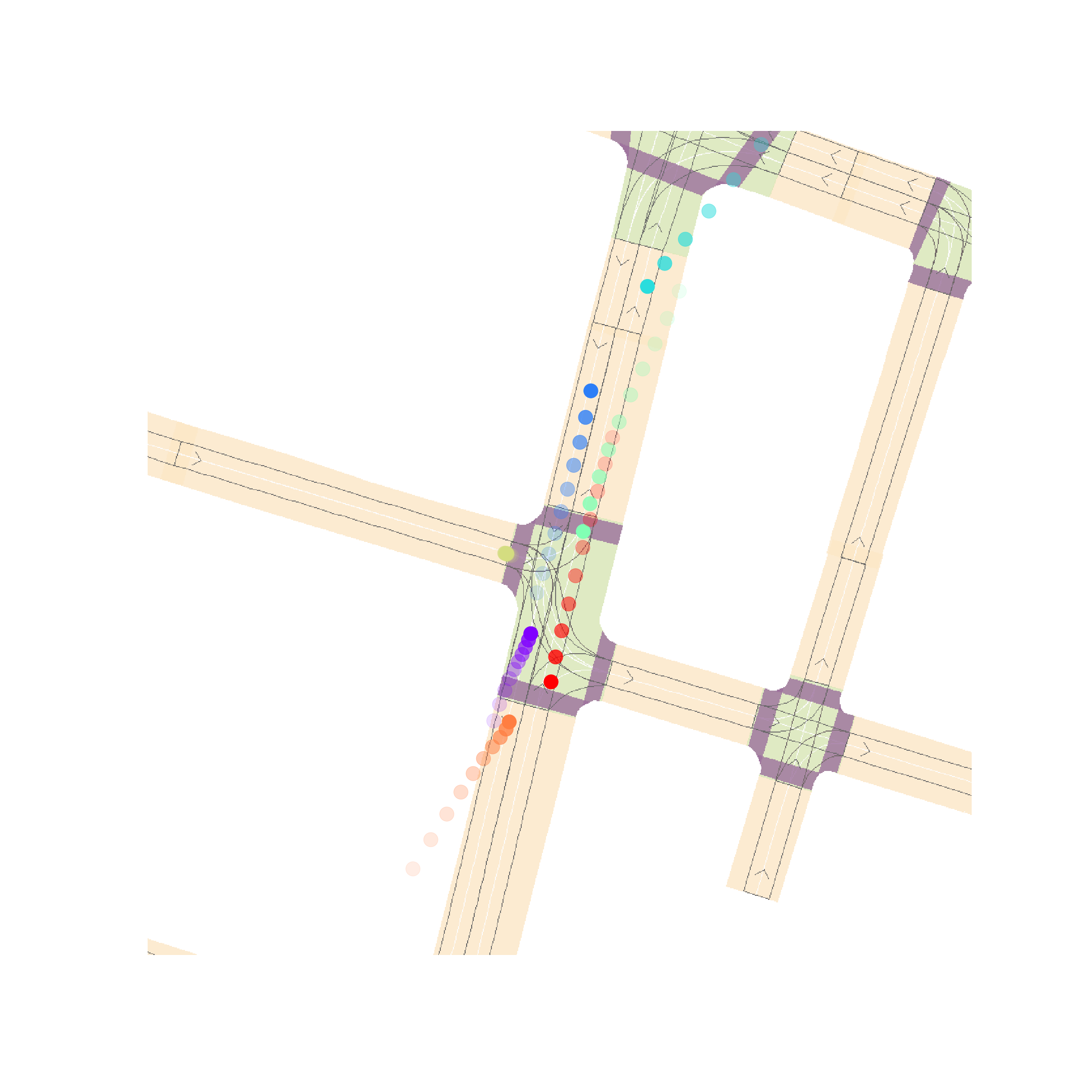}}&
      \addheight{\includegraphics[trim=300 300 300 300,clip, width=28mm]{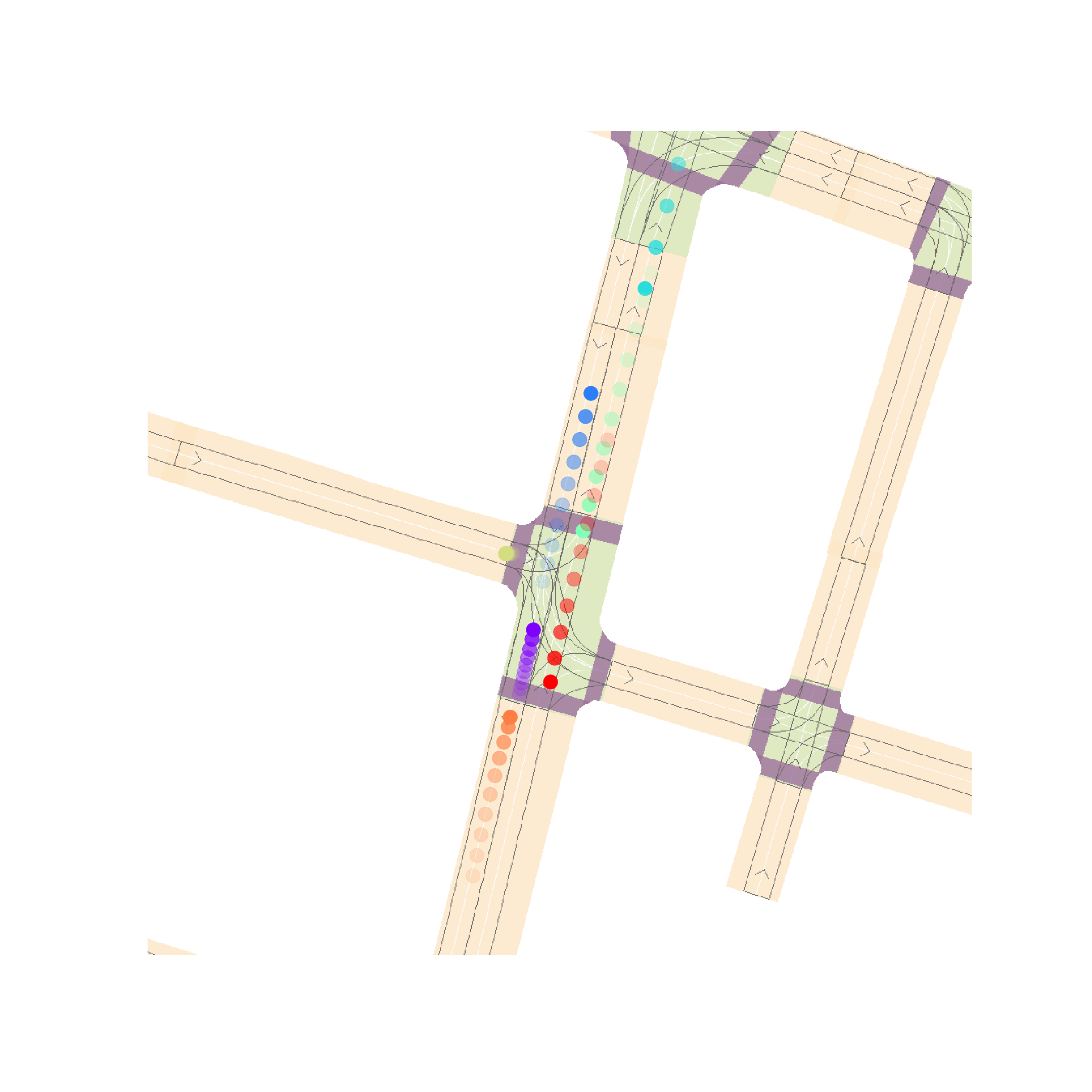}} & 
      \addheight{\includegraphics[trim=300 300 300 300,clip, width=28mm]{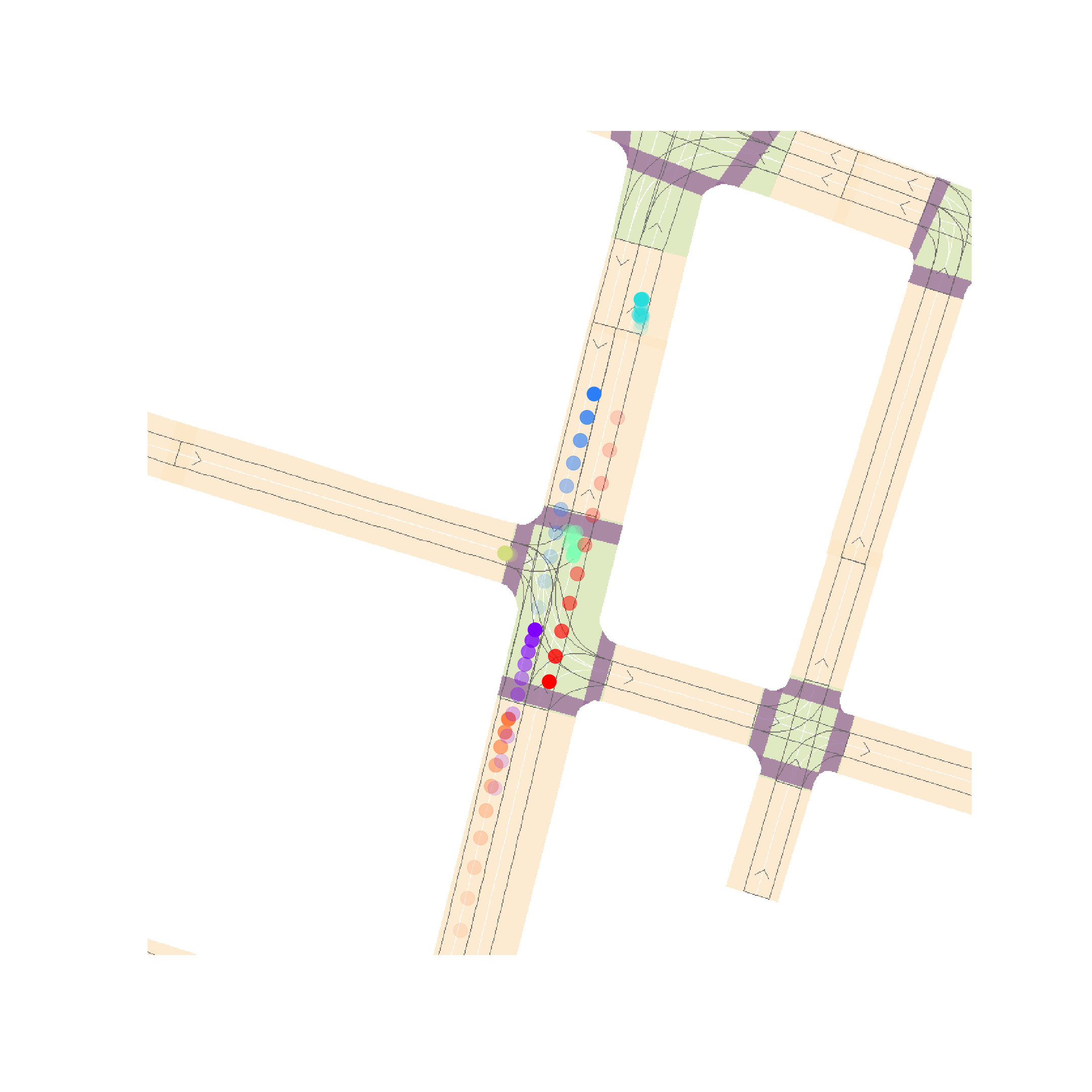}} &
      \addheight{\includegraphics[trim=300 300 300 300, clip,width=28mm]{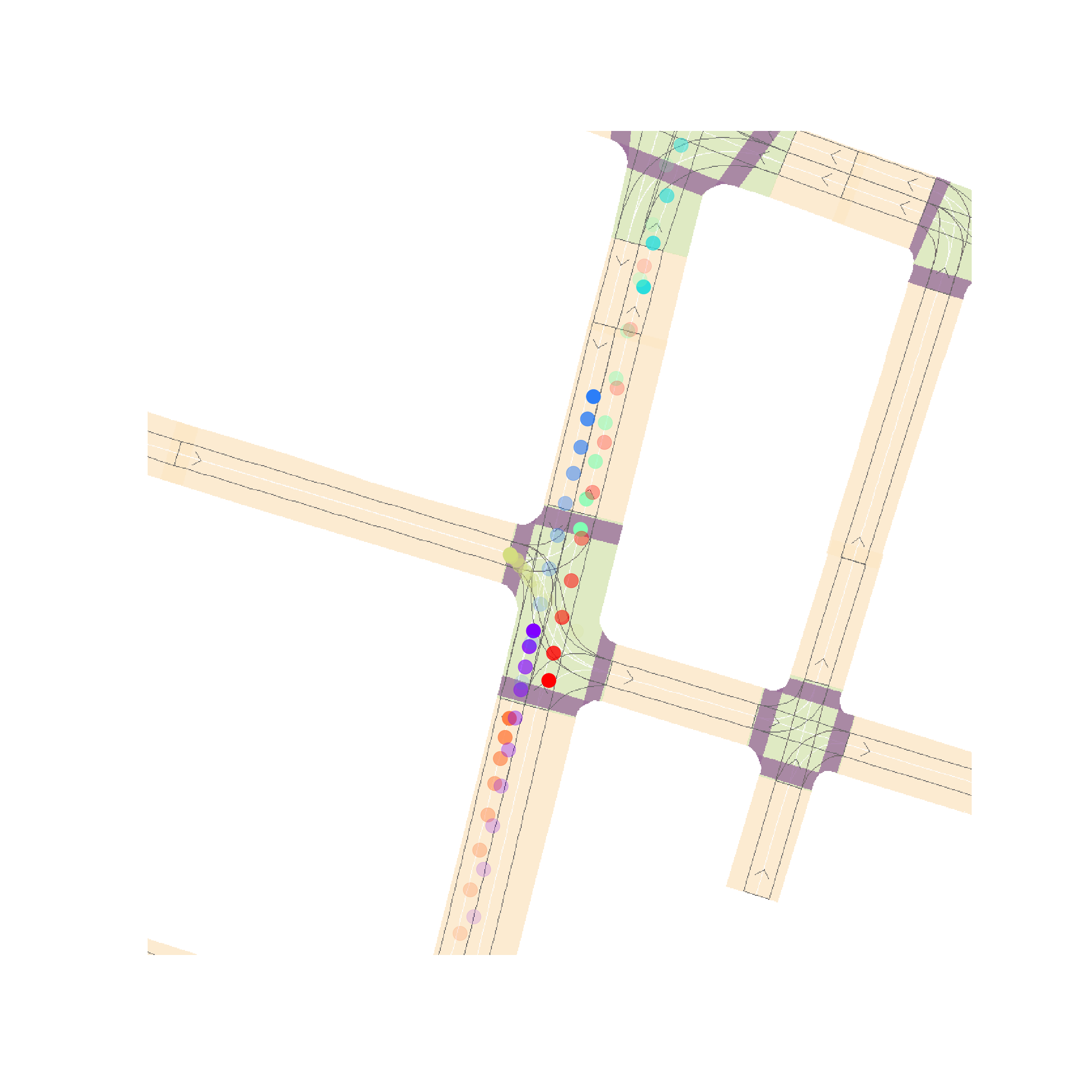}}\\      
      
     \hline
      \addheight{(c)} &
      \addheight{\includegraphics[trim=300 300 300 300, clip,width=28mm]{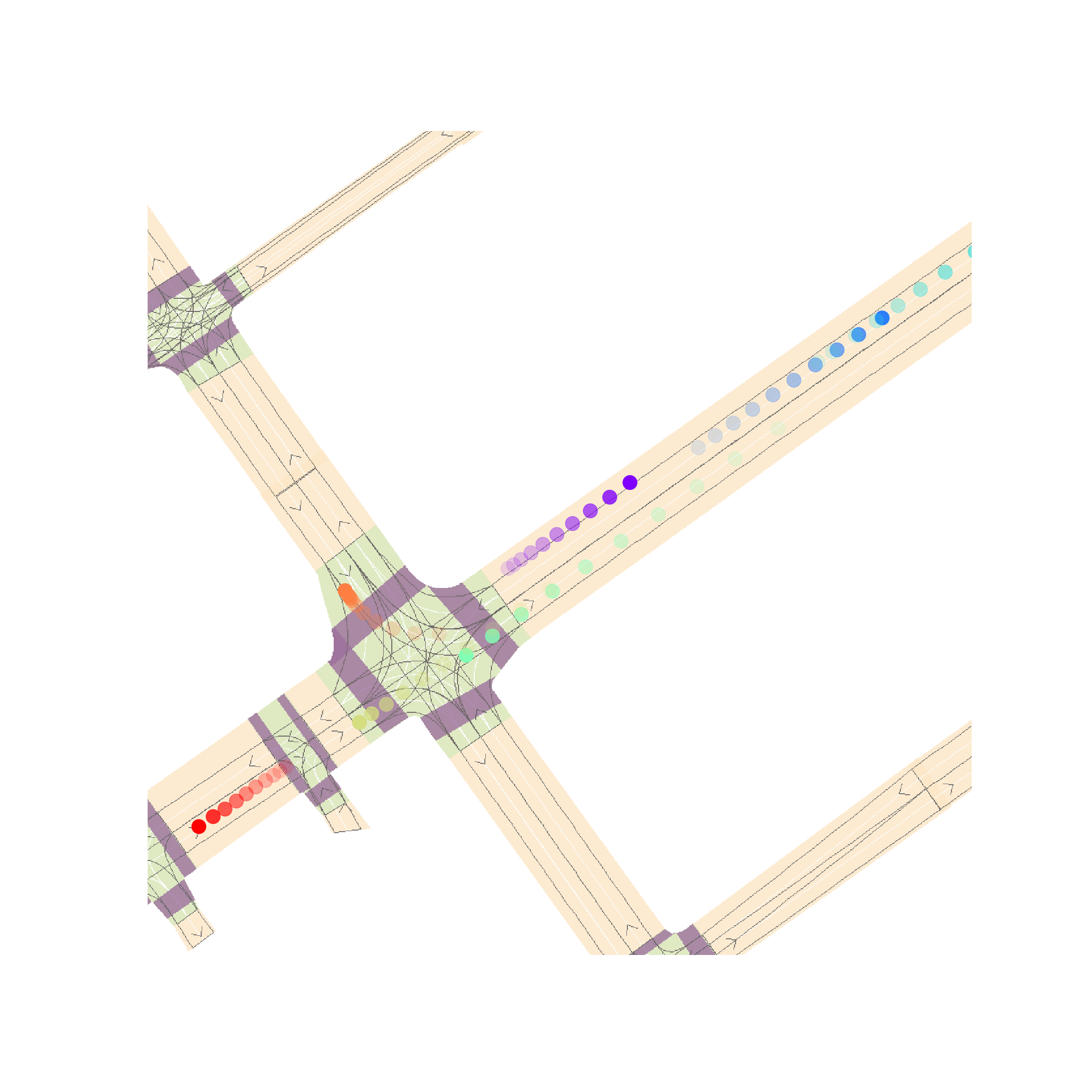}} &
      \addheight{\includegraphics[trim=300 300 300 300, clip,width=28mm]{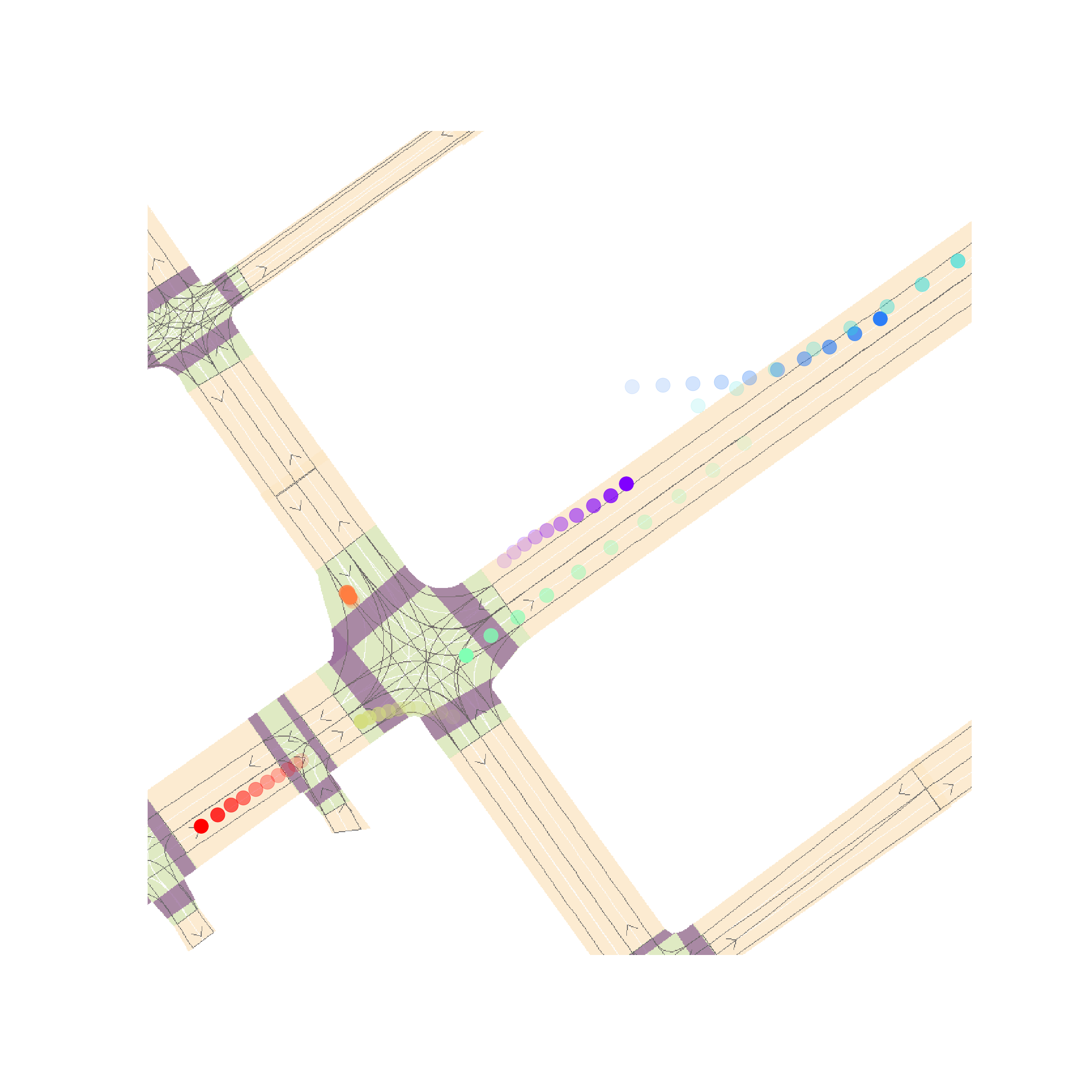}}&
      \addheight{\includegraphics[trim=300 300 300 300,clip, width=28mm]{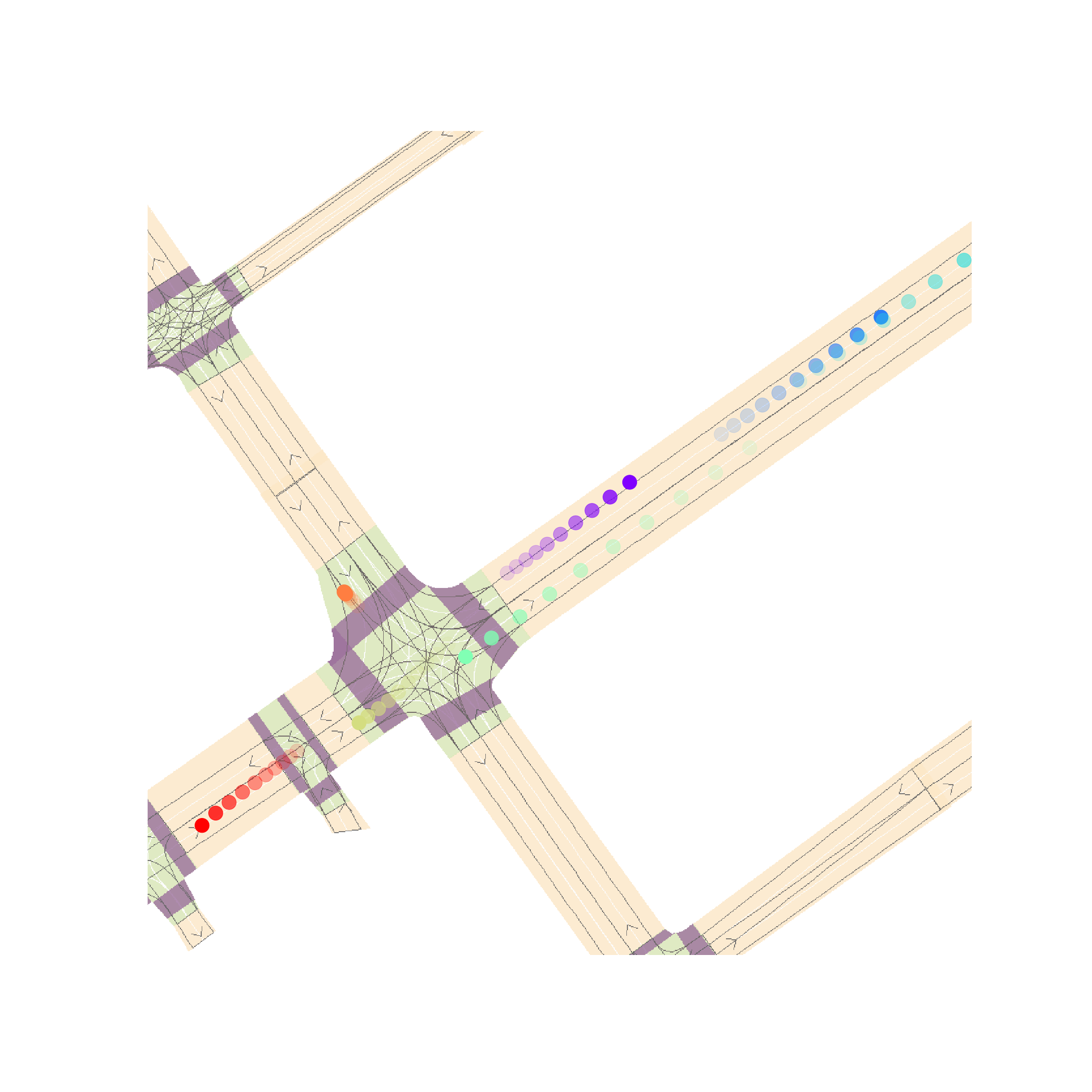}} & 
      \addheight{\includegraphics[trim=300 300 300 300,clip, width=28mm]{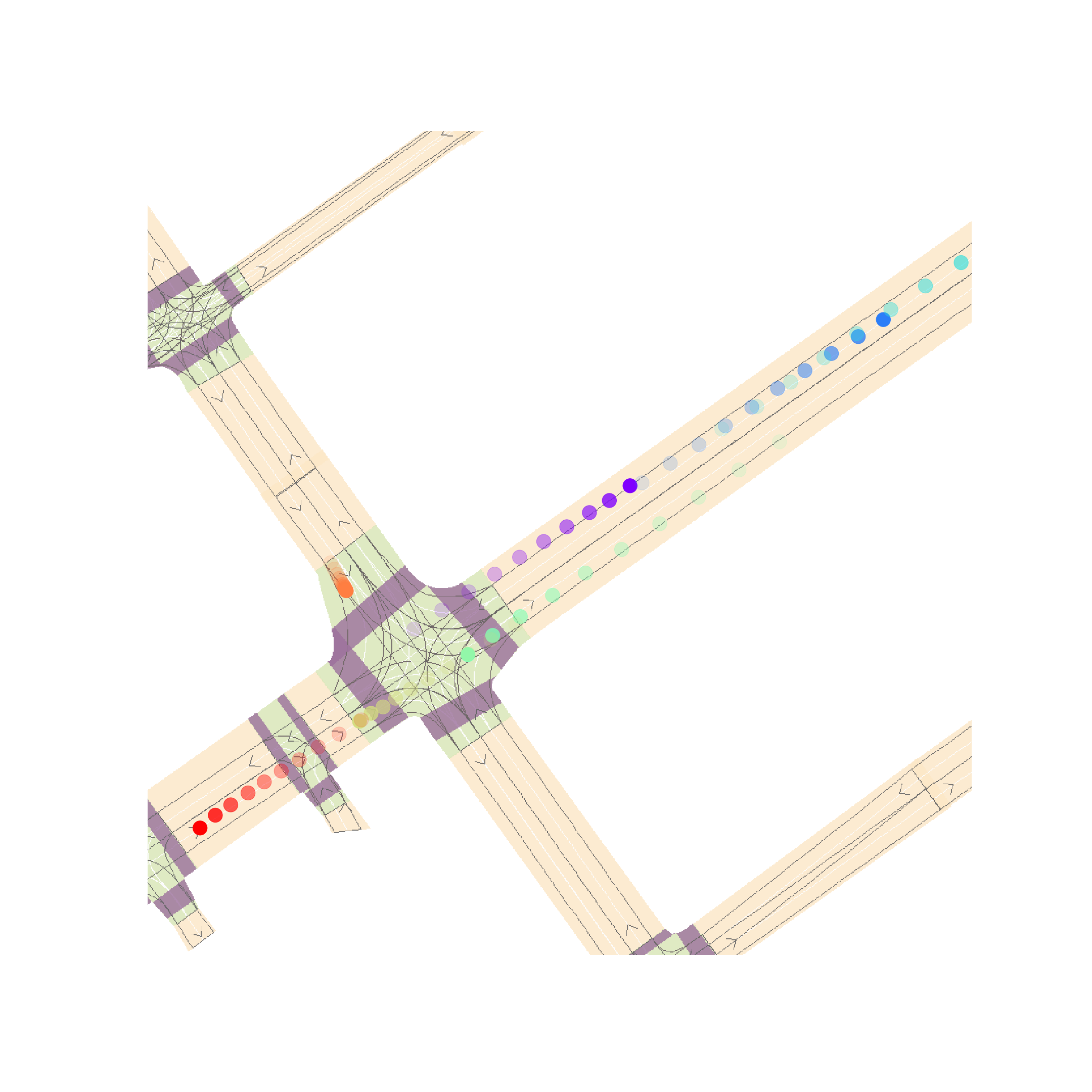}} &
      \addheight{\includegraphics[trim=300 300 300 300, clip,width=28mm]{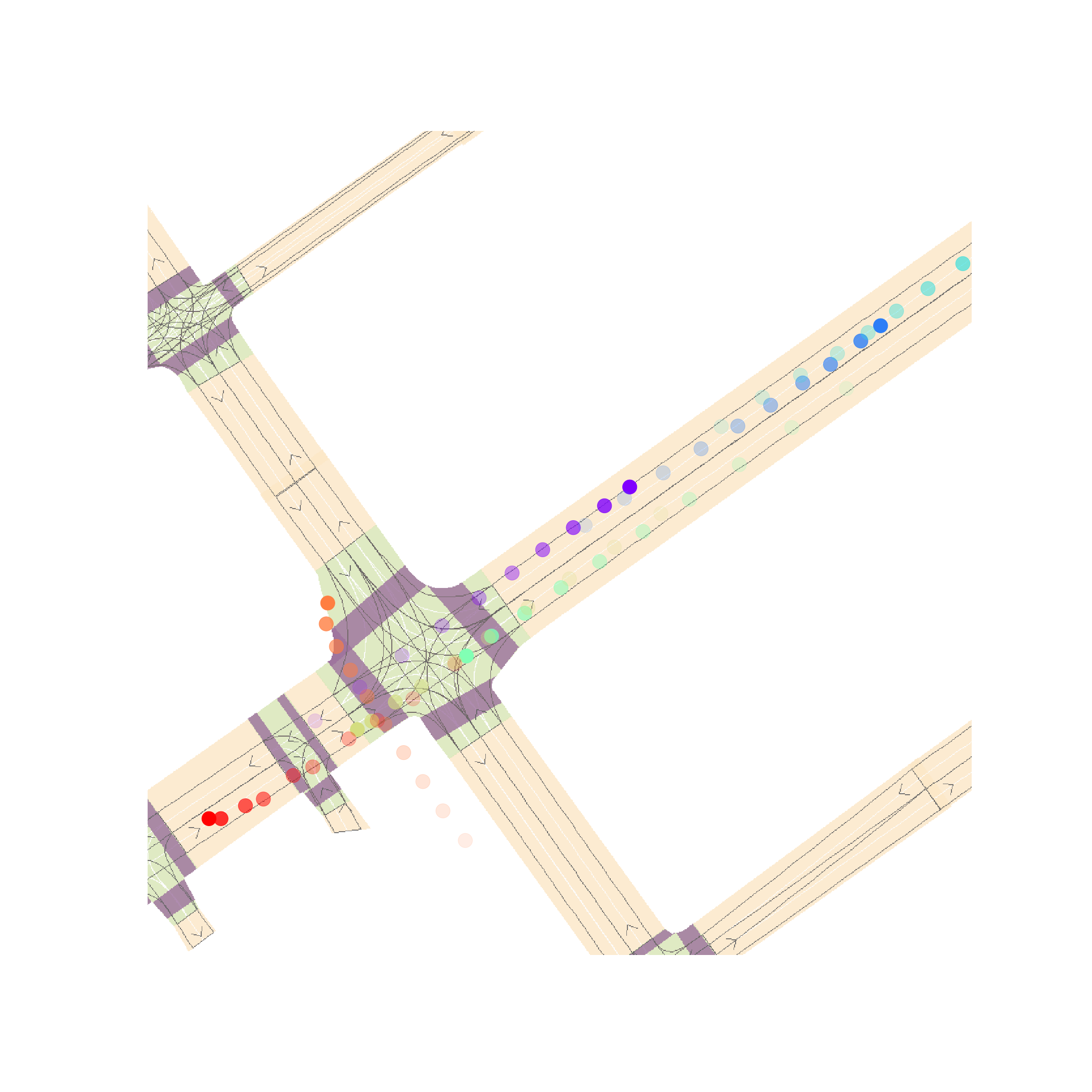}}\\
      \hline
      \addheight{(d)} &
      \addheight{\includegraphics[trim=300 300 300 300, clip,width=28mm]{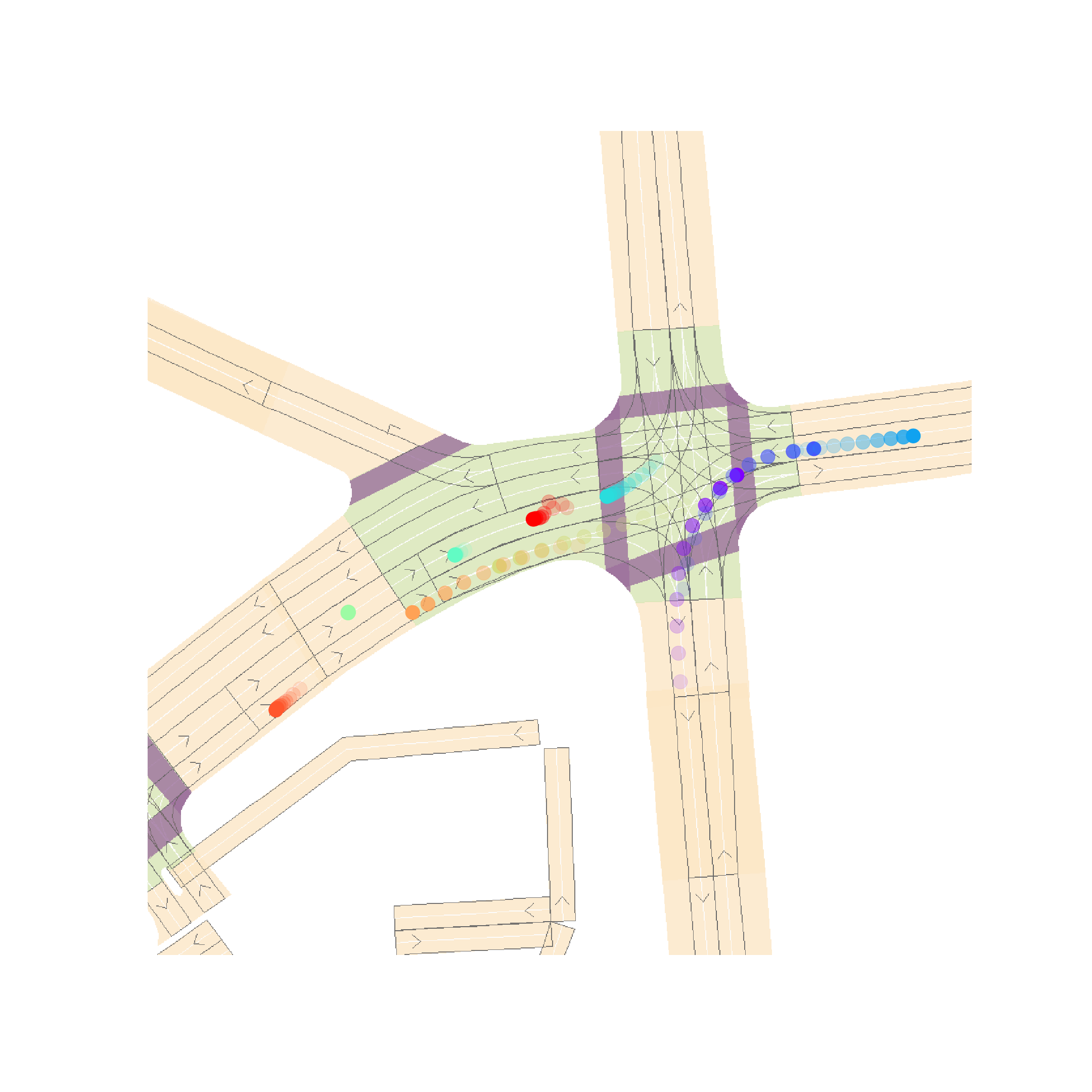}} &
      \addheight{\includegraphics[trim=300 300 300 300, clip,width=28mm]{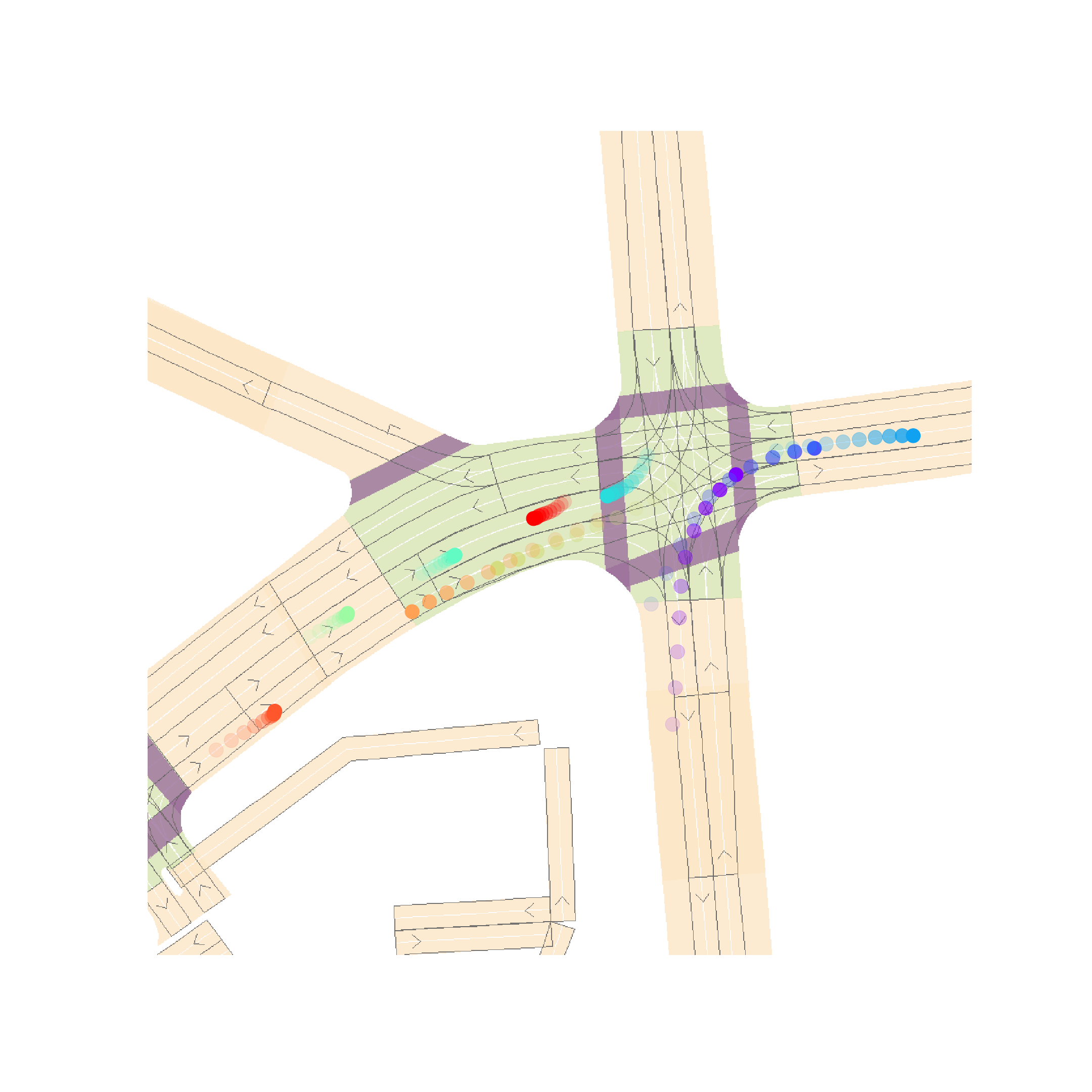}}&
      \addheight{\includegraphics[trim=300 300 300 300,clip, width=28mm]{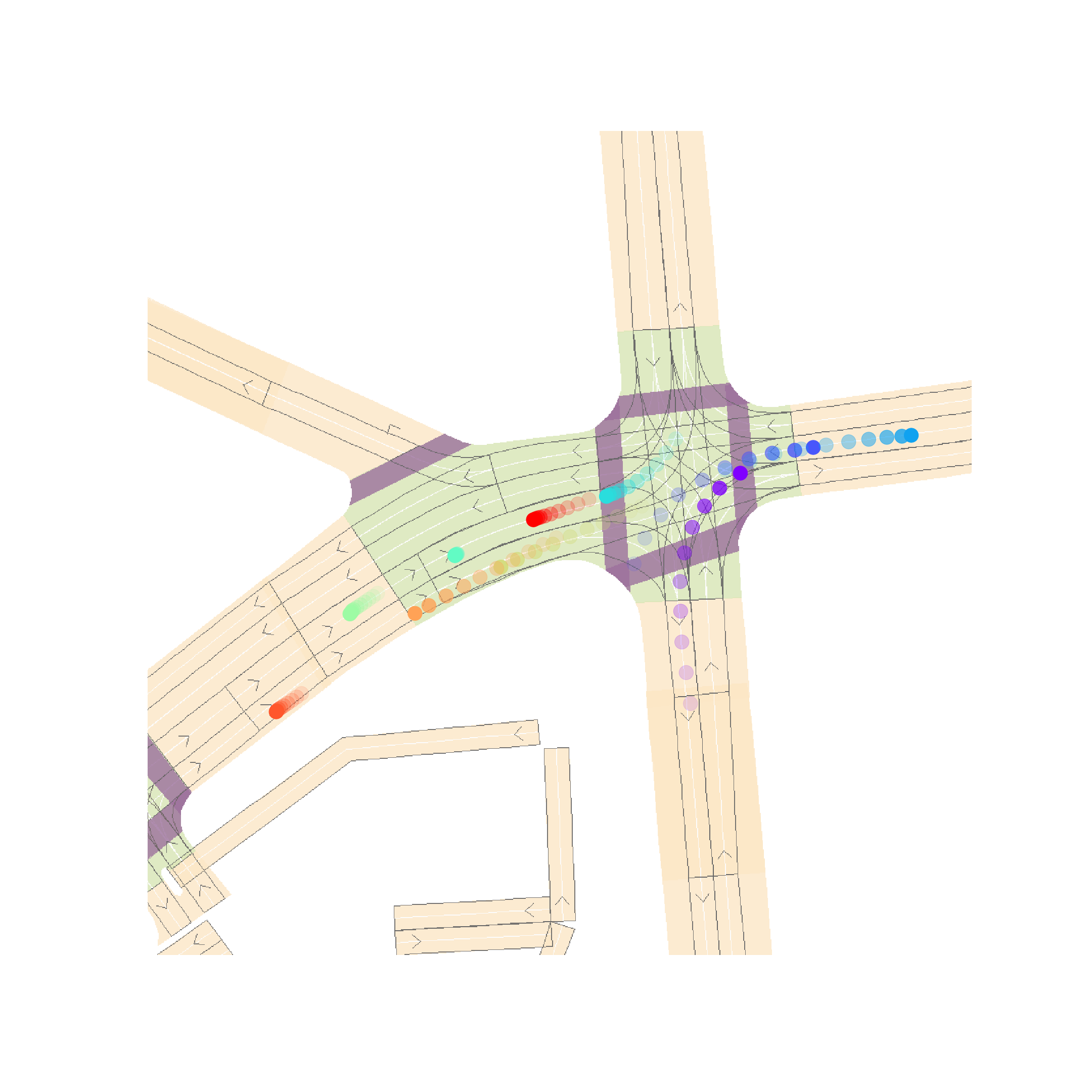}} & 
      \addheight{\includegraphics[trim=300 300 300 300,clip, width=28mm]{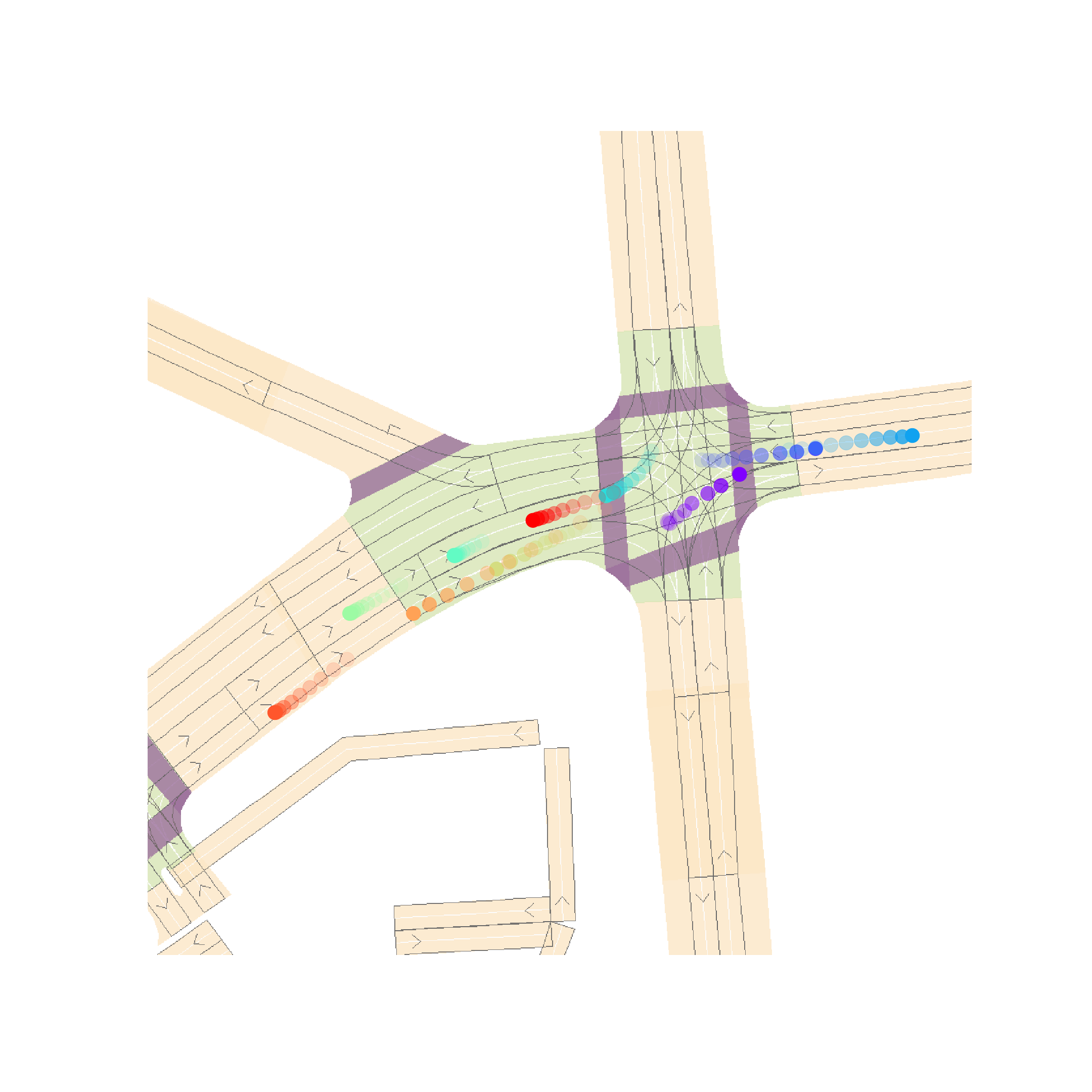}} &
      \addheight{\includegraphics[trim=300 300 300 300, clip,width=28mm]{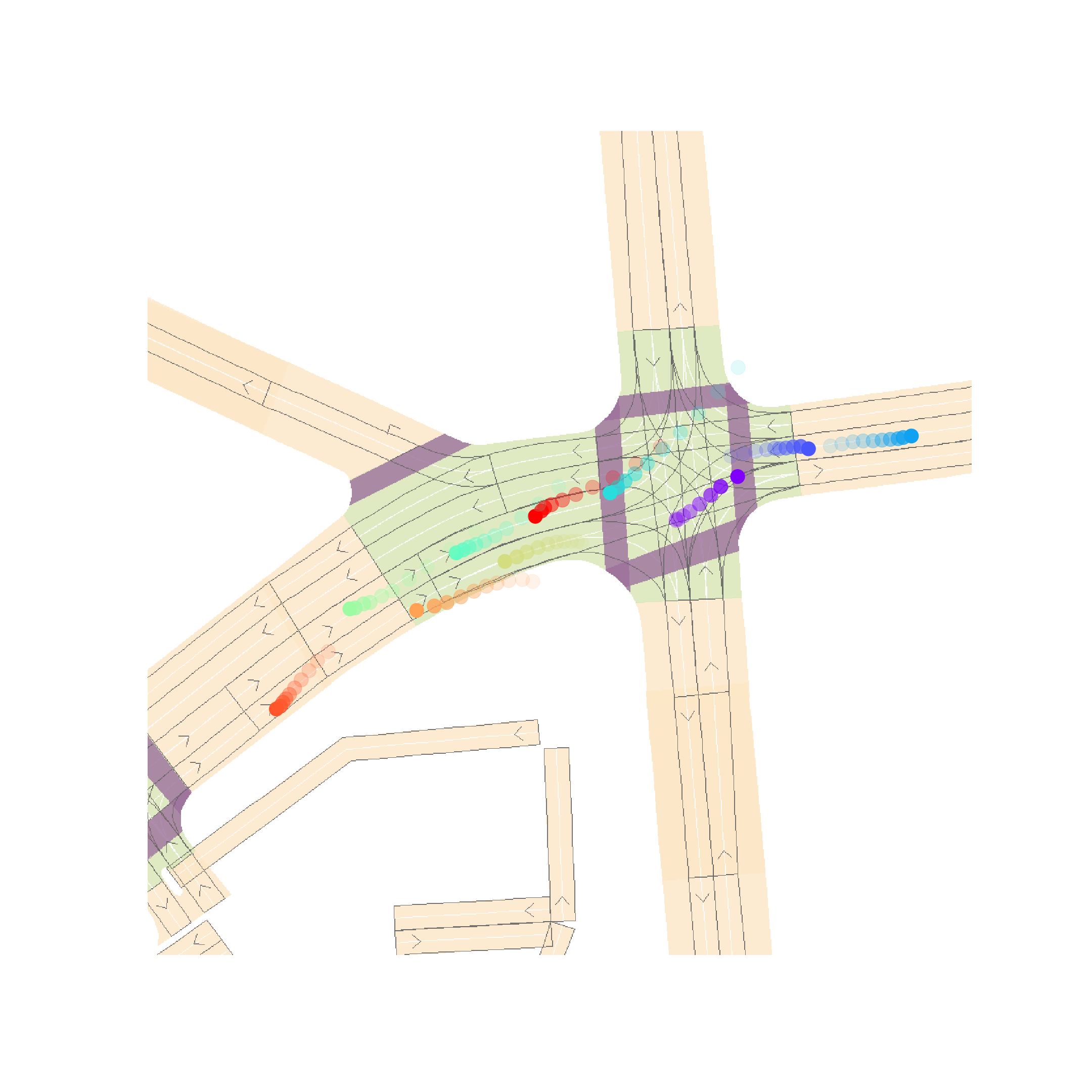}}\\      
      \hline
      \addheight{(e)} &
      \addheight{\includegraphics[trim=300 300 300 300, clip,width=28mm]{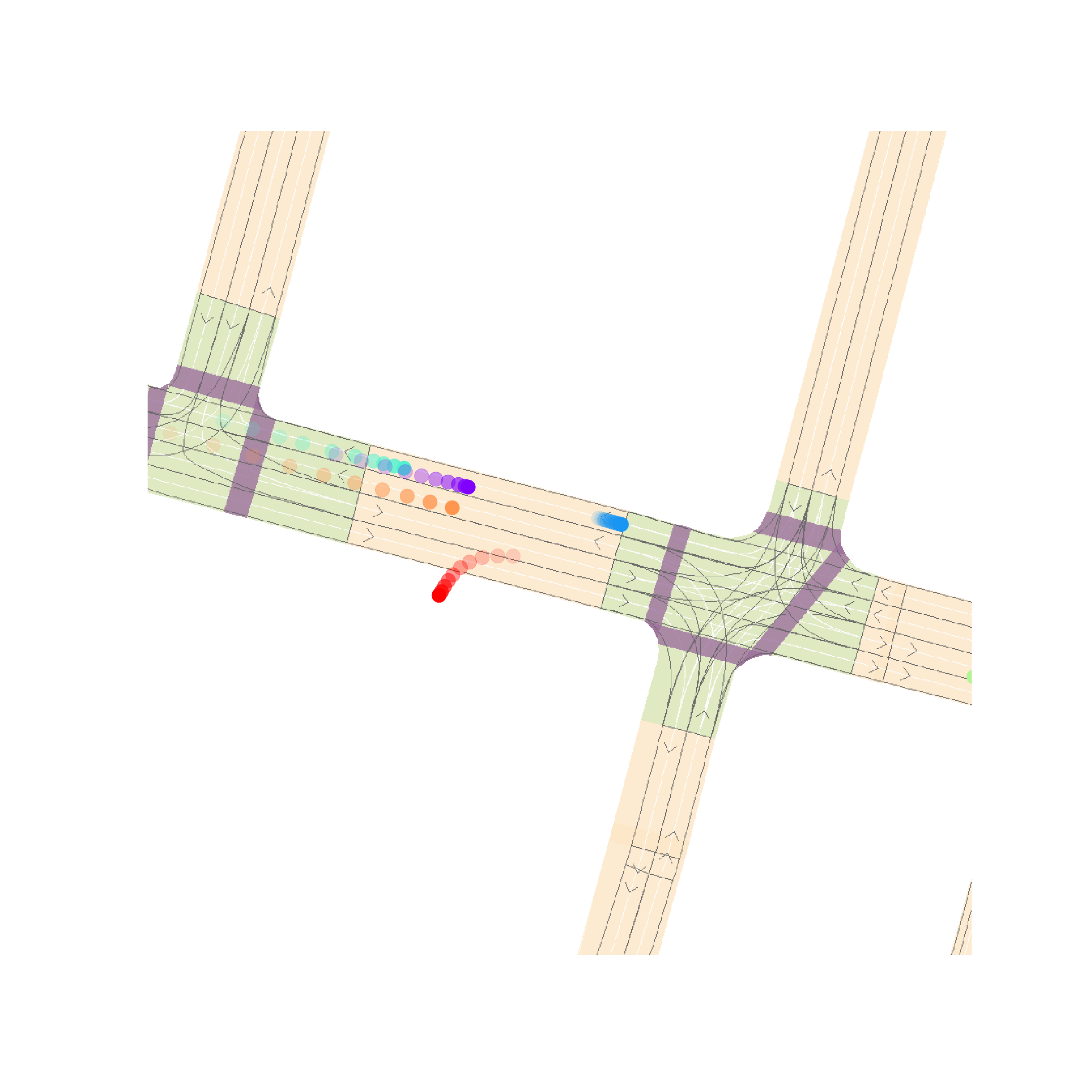}} &
      \addheight{\includegraphics[trim=300 300 300 300, clip,width=28mm]{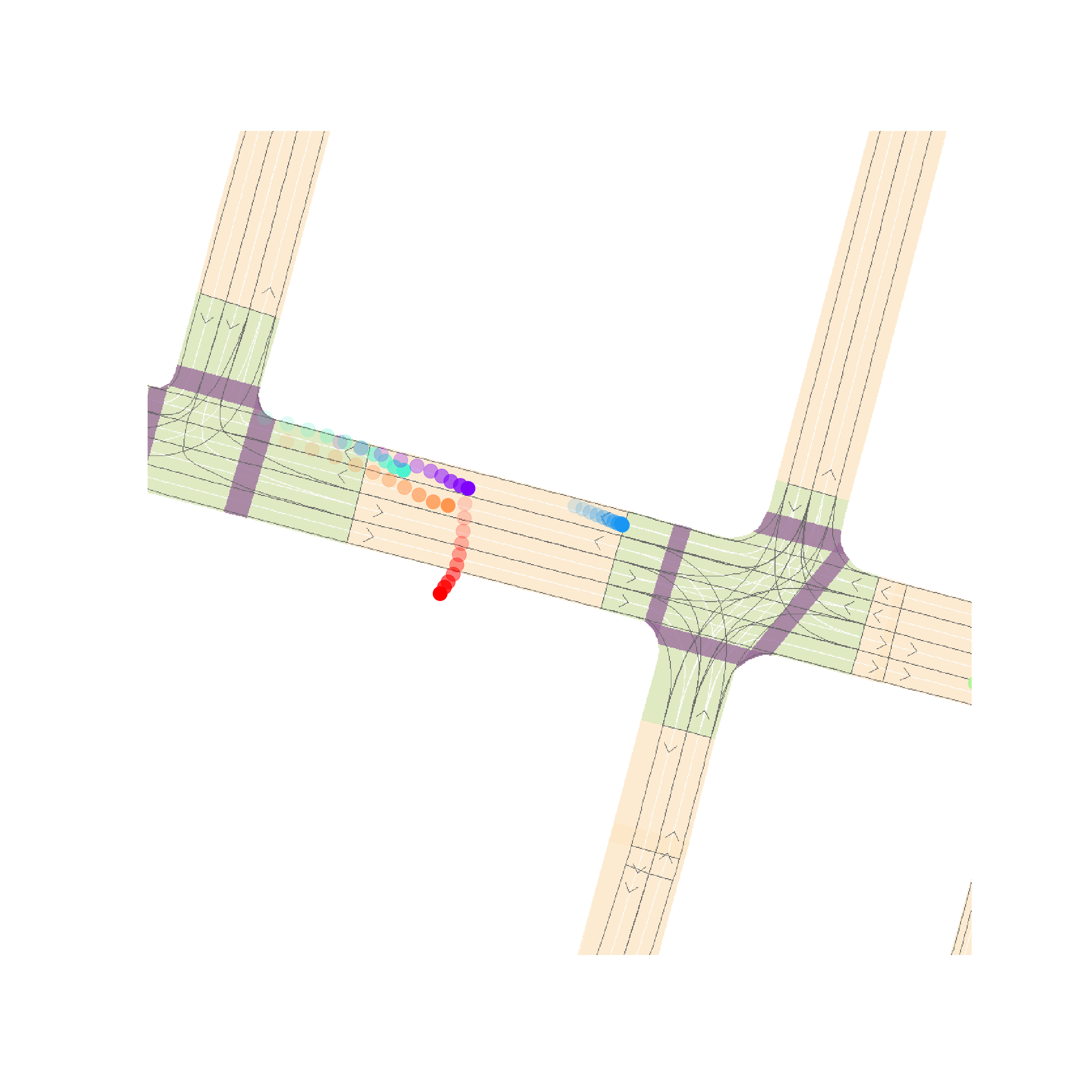}}&
      \addheight{\includegraphics[trim=300 300 300 300,clip, width=28mm]{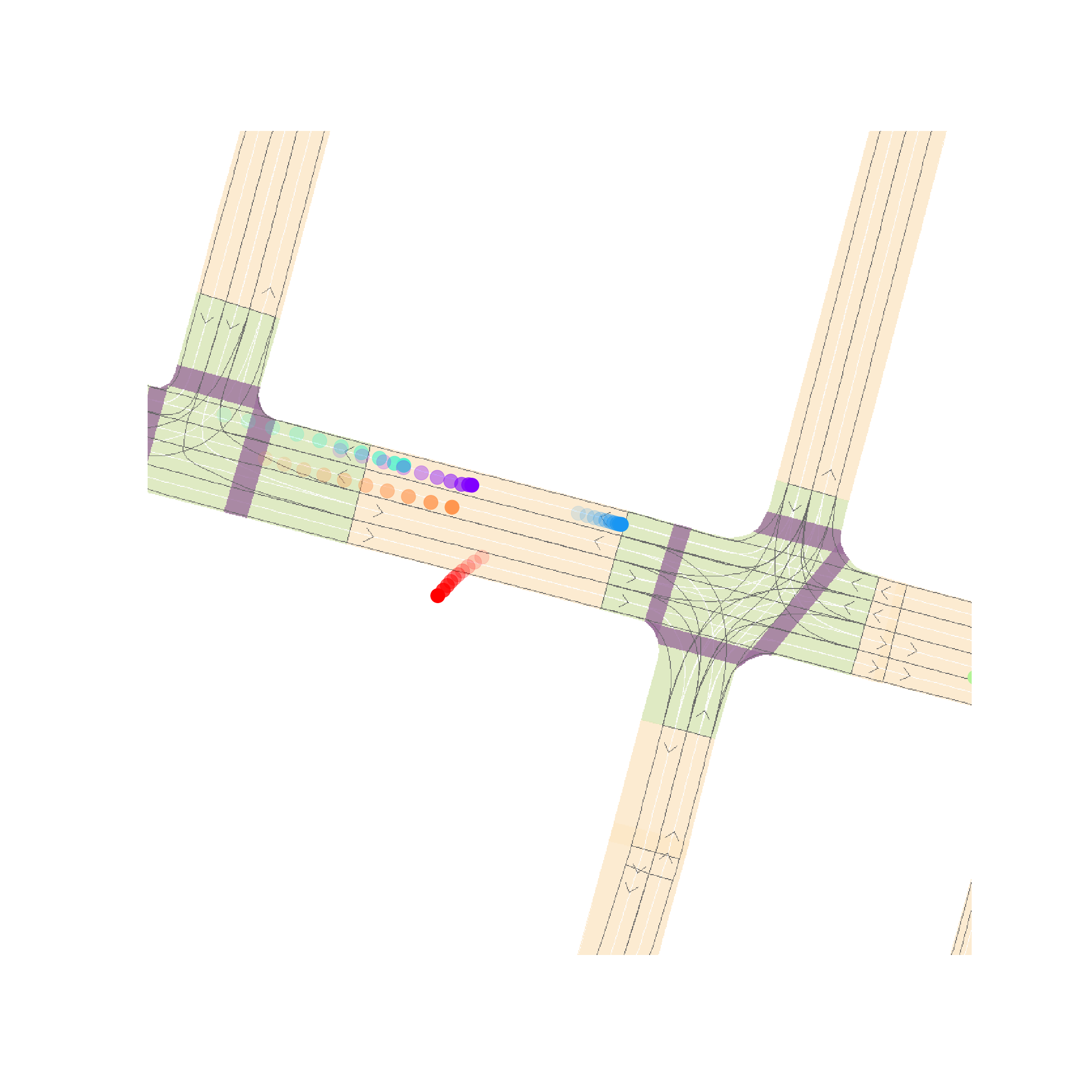}} & 
      \addheight{\includegraphics[trim=300 300 300 300,clip, width=28mm]{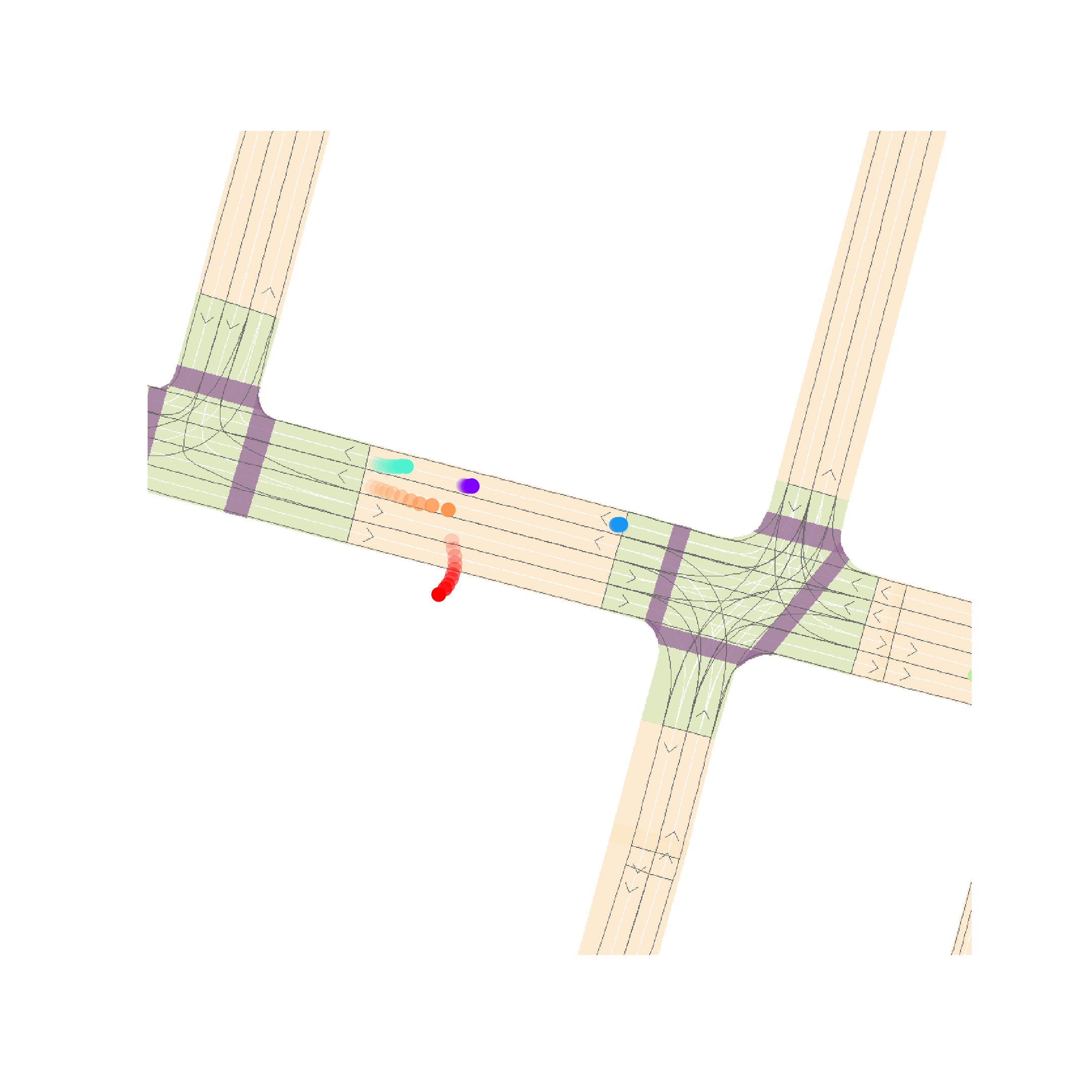}} &
      \addheight{\includegraphics[trim=300 300 300 300, clip,width=28mm]{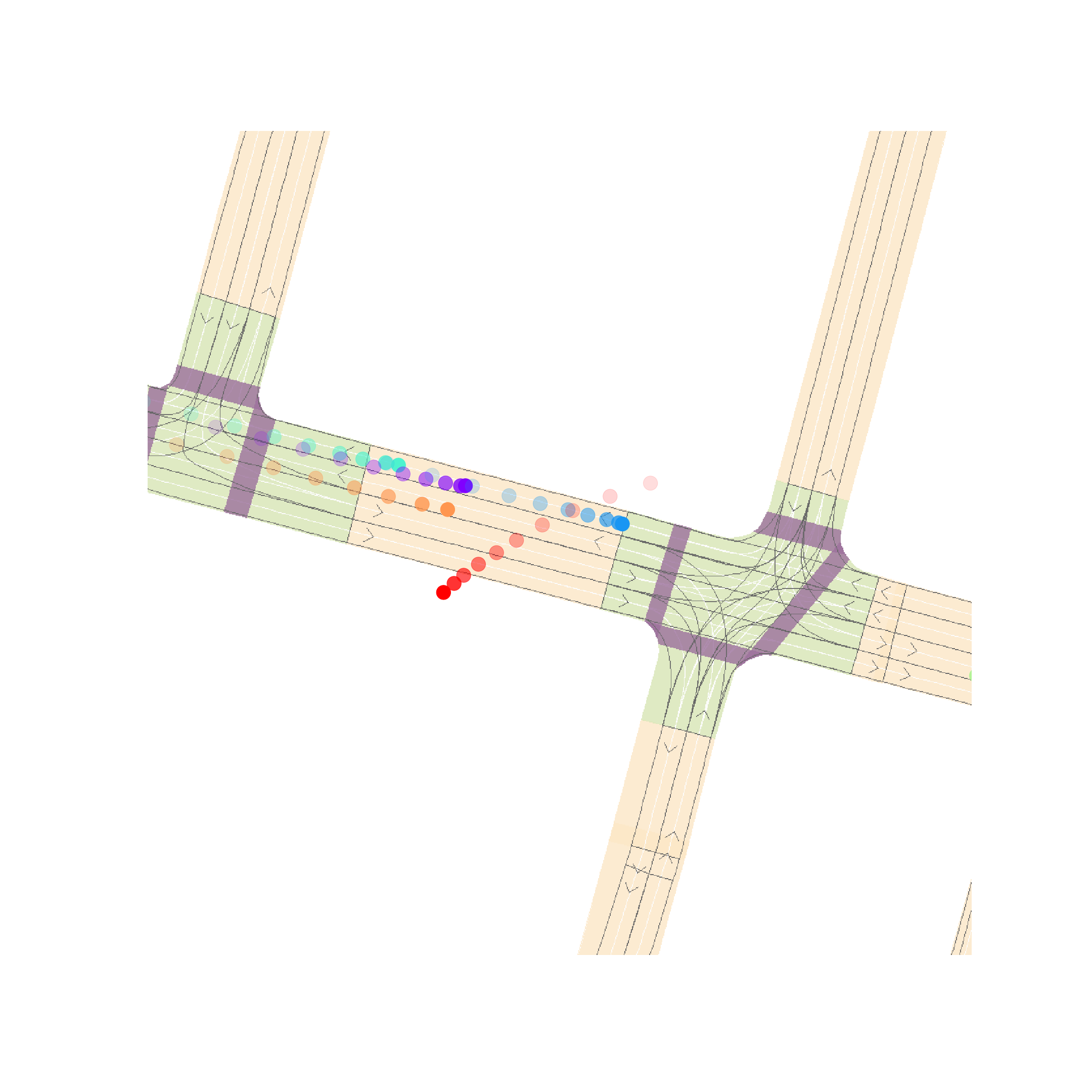}}\\      

      \hline
      \small  & ground  &  graph, oracle, & graph, oracle,& graph, joint,& graph, joint,\\      
      \small & truth &  yielding/going only &all edges & supervised  &unsupervised\\
      \hline
\end{tabular}
\vspace{.2cm}
\caption{Examples of trajectory predictions on real-world driving scenarios, from time $t$ to $t + 5s$: (a) 3-way intersection; (b) unconventional 4-way intersection; (c) canonical 4-way intersection; (d) 5-way intersection; (e) actor merging from outside the road network. The different colors indicate trajectory predictions for different actors. Each dot shows a single predicted waypoint (spaced at 0.5-second intervals), and the more transparent the dot is, the further away it is in the future (i.e.,~the further its timestamp is from the current time, $t$).}
\label{table:trajvis1}
\end{table}

\subsection{Typed Graph Network Architecture}
\begin{figure}[ht]
\centering
\includegraphics[width=.7\linewidth]{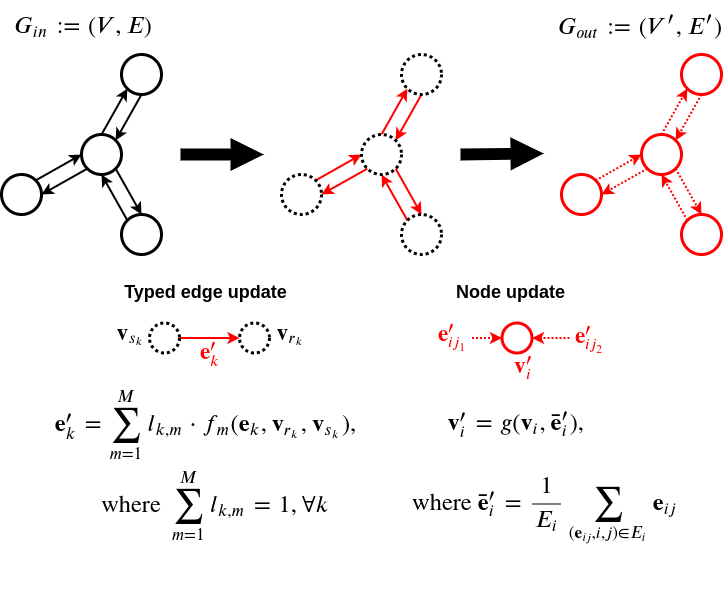}
\caption{Details of the GN layer updates with typed edges. The update function takes in $G_{in} = (V, E)$, which is an input representation for each node and each edge, and outputs $G_{out} = (V', E')$, which is the updated representation for each node and each edge. First, each edge representation is updated by processing the node attributes of its endpoints with a type-specific function $f_m$. Then, each node representation is updated by aggregating the attributes of the incoming edges.}
\label{fig:gnlayer}
\end{figure}

Here, we describe a variant of Graph Networks~\cite{battaglia2018} used in our model architecture. A Graph Network (GN) layer propagates information between the nodes and edges to output a new graph with updated representations for each node and each edge. Following the notation in \cite{battaglia2018}, a graph $G=(V, E)$ is defined by nodes $V=\{\mathbf{v_i}\}_{i=1:N^v}$ and directed edges $E=\{(\mathbf{e}_k, r_k, s_k)\}_{k=1:N^e}$, where $\mathbf{v}_i \in \mathbb{R}^{d^v}$ and $\mathbf{e}_k \in \mathbb{R}^{d^e}$ are node and edge attributes. We extend the original formulation by defining edges with discrete types and an update function for the typed edges. Assuming $M$ distinct edge types, let $l_k=(l_{k,1},...,l_{k,M})$ be the one-hot encoding or the scores of the types for edge $k$. Then, the typed edge update function outputs updated edge attribute $\mathbf{e}_k' = \sum_{m=1}^{M} l_{k,m} \cdot f_m(\mathbf{e}_k, \mathbf{v}_{r_k}, \mathbf{v}_{s_k})$, where $f_m$ is a learnable function for each edge type $m$. Additional details can be found in Figure~\ref{fig:gnlayer}.

\subsection{Additional Quantitative Comparison}

\begin{figure}
\begin{subfigure}{.33\linewidth}
\centering
\includegraphics[width=\linewidth]{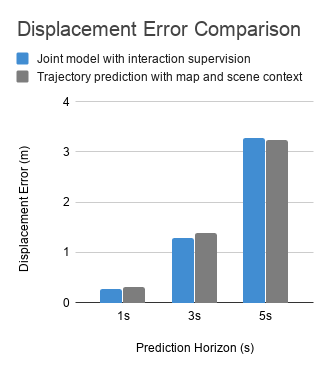}
\label{fig:sub1}
\end{subfigure}%
\hfill
\begin{subfigure}{.33\linewidth}
\centering
\includegraphics[width=\linewidth]{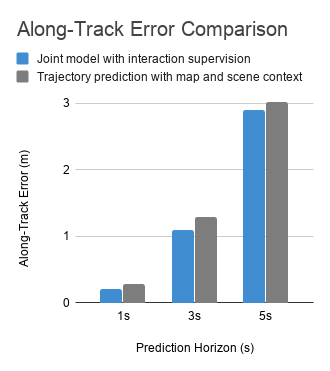}
\label{fig:sub2}
\end{subfigure}
\hfill
\centering
\begin{subfigure}{0.33\linewidth}
\centering
\includegraphics[width=\linewidth]{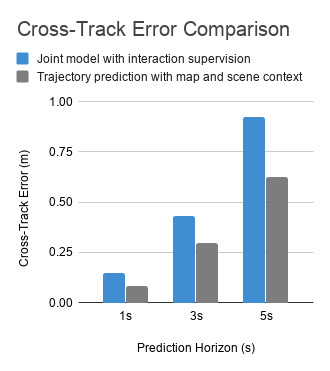}
\label{fig:sub3}
\end{subfigure}
\caption{A comparison of the trajectory prediction error between our proposed supervised joint model (shown in blue) and the baseline from~\cite{djuric2018motion} (shown in gray). From left to right, the plots show the displacement error, along-track error, and cross-track error at multiple different time horizons. All errors are reported in meters. We see that our approach is comparable in terms of displacement error, better in terms of along-track error, and worse in terms of cross-track error.}
\label{fig:quanti_analy}
\end{figure}

In Figure~\ref{fig:quanti_analy}, we further compare our supervised joint model to the baseline from \cite{djuric2018motion} by measuring the trajectory errors at different time horizons (1 second, 3 seconds, 5 seconds). 
The results indicate that our approach is worse than the baseline in terms of cross-track error, which is expected because \cite{djuric2018motion} provides a rasterized bird's eye view of the map as an input to the model, and we don't have the same map and scene context.  
However, we also see that our method is better than the baseline in terms of along-track error, which highlights the value of explicitly capturing temporal interactions such as going and yielding for the trajectory prediction problem.

\medskip

\small

\end{document}